\documentclass[11pt]{article}

\usepackage[utf8]{inputenc}
\usepackage{graphicx}
\usepackage{fancyhdr}
\usepackage{amsmath}
\usepackage{apacite}
\usepackage{array}
\usepackage{booktabs}
\newcolumntype{P}[1]{>{\centering\arraybackslash}p{#1}}
\usepackage{placeins}
\usepackage{multirow}
\usepackage{array}
\usepackage{setspace}
\title{\Huge Finding the Answers \\ With Definition Models}
\date{}
\usepackage[margin={1in,1in}]{geometry}
\usepackage{titling}

\author{Jack Parry}
\begin{document}

\pagenumbering{gobble}

\maketitle

\begin{center}
Word Count: 9603

\includegraphics{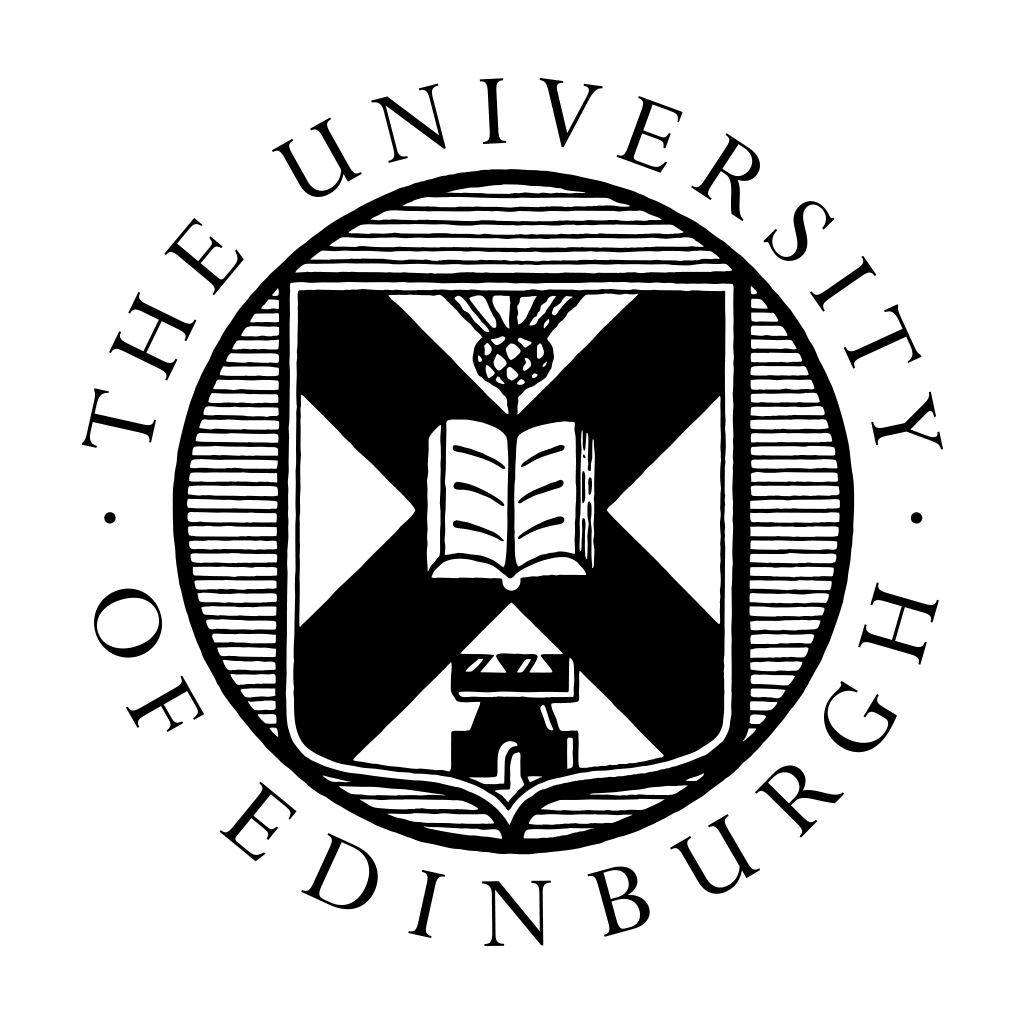}

\small MSc Dissertation

Speech and Language Processing

School of Philosophy, Psychology, Languages and Linguistics

University of Edinburgh

2018
\end{center}

\newpage
\normalsize
\pagenumbering{roman}

\begin{center}
\section*{Acknowledgements}
I would like to thank my friends and fellow coursemates for being a constant source of support, guidance and motivation. I extend my gratitude to Felix Hill for promptly supplying the code and data, without which much of this work would not have been possible. I also thank my supervisor Antonio Miceli Barone for his helpful comments and technical expertise. 
\end{center}

\newpage
\begin{spacing}{1}

\begin{abstract} \normalsize Inspired by a previous attempt to answer crossword questions using neural networks \cite{hill2015learning}, this dissertation implements extensions to improve the performance of this existing definition model on the task of answering crossword questions. A discussion and evaluation of the original implementation finds that there are some ways in which the recurrent neural model could be extended. Insights from related fields neural language modeling and neural machine translation provide the justification and means required for these extensions. Two extensions are applied to the LSTM encoder, first taking the average of LSTM states across the sequence and secondly using a bidirectional LSTM, both implementations serve to improve model performance on a definitions and crossword test set. In order to improve performance on crossword questions, the training data is increased to include crossword questions and answers, and this serves to improve results on definitions as well as crossword questions. The final experiments are conducted using sub-word unit segmentation, first on the source side and then later preliminary experimentation is conducted to facilitate character-level output. Initially, an exact reproduction of the baseline results proves unsuccessful. Despite this, the extensions improve performance, allowing the definition model to surpass the performance of the recurrent neural network variants of the previous work \cite{hill2015learning}.

\end{abstract}
\end{spacing}
\newpage
\normalsize
\tableofcontents

\newpage
\begin{spacing}{1}
\normalsize
\pagenumbering{arabic}
\section{Introduction}
This dissertation directly builds upon the model implemented by Hill et al. \citeyear{hill2015learning}. The model is defined here as a `definition model', and it is closely related to a language model. The difference is that while a language model takes a sequence of words and tries to predict the following word, a definition model takes a definition of a word and tries to predict the word that is being described. These models are also often described as `reverse dictionary models'. Definition models have been successfully developed in the past \cite{bilac2004dictionary}, but attempts at doing so are not great in number. Hill et al.'s \citeyear{hill2015learning} paper marks one of the first attempts at creating a neural definition model, following on from recent research in natural language processing. This dissertation contributes to research on definition models, applying state-of-the-art techniques from the fields of neural language modeling and neural machine translation to Hill et al.'s \citeyear{hill2015learning} definition model to improve performance on the available test sets. There is a focus on improving the model's performance with answering general knowledge crossword questions in order to make the case that these models can learn some degree of inference or reasoning, something often thought to be very difficult for neural networks \cite{NIPS2017_7082}. In order to facilitate this, crossword questions and answers have been gathered to add to the training data for these models.

This dissertation begins by reviewing the work of Hill et al. \citeyear{hill2015learning}, then identifies the specific ways in which we will extend the model in 2.5. Section 3 introduces some recent contributions to the fields of neural language modeling and neural machine translation. Each of these contributions can be applied to definition models, and are explained before we outline how they are implemented and why. Section 4 introduces the model developed by this dissertation, chiefly explaining how this model differs from the one developed by Hill et al \citeyear{hill2015learning}. Section 5 includes the results and discussion of all of the main experiments, with model extensions outlined and evaluated in 5.4 and 5.5. Section 6 discusses the potential benefits of adding a fully character-level decoder to the model, and qualitatively evaluates some samples from a baseline character-level model trained on Nematus \cite{sennrich-EtAl:2017:EACLDemo}.

\section{Previous Work: Hill et al. (2015)}
Recent NLP research has been hugely successful in using distributional models to learn representations of individual words \cite{Turian:2010:WRS:1858681.1858721}, \cite{mikolov2013efficient}, but it has proven more difficult to create representations of phrases. The motivation for Hill et al.'s \citeyear{hill2015learning} work was to develop a model that could successfully compose the meaning of phrases and sentences that describe a word into a vector representation of the word. Such models are henceforth referred to as `definition models'. These models are similar to language models, but rather than predicting the next word given the previous words in a sequence, they predict the head word given the gloss words of a dictionary definition. 

The motivation for their work comes from the potential real-world applications of such a model. One such application is as a tool for journalists or authors to search for words by entering a description. A second use is to answer questions, such as general knowledge crossword questions. The existence of commercial systems designed to perform these functions (reverse dictionary model: \textit{OneLook}\footnote{https://www.onelook.com/reverse-dictionary.shtml}; crossword solving model: \textit{crossword maestro}\footnote{http://www.crosswordmaestro.com/}) shows that there is a demand for such systems. The paper explores various ways of mapping a representation of a dictionary definition (henceforth `gloss') to an embedding of the word it describes (henceforth `head'). The success of the authors in producing a model that gives comparable or better results than some commercial task-specific systems for their chosen evaluation metrics shows that they have created model that can to some extent capture the meanings of phrases and sentences.

\subsection{Data}
Data gathered by Hill et al. \citeyear{hill2015learning} in preparation for the experiments includes a training and test set and a set of pretrained word embeddings. The pretrained embeddings were trained on very large corpora (8bn words of running text) using Word2Vec (W2V)\footnote{https://code.google.com/archive/p/word2vec/} \cite{mikolov2013efficient}. An explanation of W2V software is included in Section 2.4. These pretrained embeddings are used as the representation for the head words in all experiments, meaning that this is the target to which the composed gloss representation is compared. While the gloss embeddings are usually learned, the authors also experiment with using pretrained embeddings for the gloss words in the hope that these embeddings may capture a richer semantic representation of the word because of their larger training corpora.

The set of words for which definitions were gathered (as training and test data) were the unique words in the target embedding space. Word definition pairs were extracted from several dictionaries plus the words' respective Wikipedia articles. The dictionaries were chosen due to their open domain nature and accessibility via the WordNik API\footnote{http://developer.wordnik.com} \cite{hill2015learning}. For the Wikipedia data, if the word in question has a Wikipedia page, each sentence in the first paragraph is considered a separate gloss for that word. Hill et al \citeyear{hill2015learning} argue that this should help the system access more facts about the target words, although this dissertation believes it likely introduces some less accurate definitions of the words into the training data, because not each sentence in Wikipedia will necessarily accurately describe the head word. Most words have more than one definition across these media, in which case, all definitions are included in the training data. The training data is made up of $\sim850,000$ head gloss pairs, with $\sim90,000$ unique head words.

For model evaluation, four test sets are created \cite{hill2015learning}: 
\begin{itemize}
\item Seen set (500 word definition pairs taken from training data)
\item Unseen set (500 dictionary definitions for unseen head words)
\item Concept descriptions (200 user generated definitions for head words in vocab)
\item Crossword set (200 ``general knowledge'' crossword questions taken from a crossword website\footnote{www.eddiejames.com, now unavailable})
\end{itemize}

While the authors experiment with a subsection of the training set (only WordNet definitions), they find that the best performance on all tasks is achieved on the full training set. The smaller WordNet training set however consists of only 150,000 word definition pairs, which is a dramatic reduction in size ($<20\%$ of the full training set). It would have been interesting to see more experimentation with the training data, particularly excluding the Wikipedia data which may have introduced some inaccurate definitions into the data, as mentioned above. While some experiments conducted in this dissertation will add crossword data to the training set, specific original sources cannot be removed from the data because the training examples included in the repository are not labeled with their source.

\subsection{Models}
Hill et al. \citeyear{hill2015learning} implement several models:

Baseline models:
\begin{itemize}
\item W2V add
\item W2V mult
\end{itemize}

Neural models:
\begin{itemize}
\item Recurrent neural network (RNN)
\item Bag-of-words (BoW)
\end{itemize}
The baseline models are based on Mitchell and Lapata's work \citeyear{mitchell2010composition}, they work by combining the individual gloss word embeddings by either pointwise addition (W2V add) or elementwise multiplication (W2V mult). The resulting embedding is compared with all embeddings in the vocabulary using cosine similarity to rank the candidates. Mitchell and Lapata \citeyear{mitchell2010composition} explain that the multiplicative model can only affect the magnitude of the resultant vector and not the direction. Given that cosine similarity is insensitive to differences in magnitude this could be the reason that the multiplicative model proves less effective than the additive model as shown in Table \ref{hill_results_1}.

\begin{table}[!htbp]
\caption[ ]{Dictionary definition results in Hill et al. (2015)}
\begin{tabular}{ P{1cm} p{3cm} | p{0.5cm} p{1cm} p{0.5cm} | p{0.5cm} p{1.2cm} p{0.5cm} |  p{0.5cm} p{1cm} p{0.5cm} }
\toprule
\multicolumn{2}{p{2cm} |}{Test Set}                     & \multicolumn{3}{p{3cm} | }{Seen (500 WN defs)}      & \multicolumn{3}{p{3.5cm} | }{Unseen (500 WN defs)}      & \multicolumn{3}{p{4.5cm}}{Concept descriptions (200)}  \\\midrule
\multirow{2}{1cm}{Unsup. models} & W2V add         & -          & -                & -           & 923         & .04/.16           & 163         & 339           & .07/.30         & 150           \\
                               & W2V mult        & -          & -                & -           & 1000        & .00/.00           & 10*         & 1000          & .00/.00         & 27*           \\\hline
                               & OneLook         & \textbf{0} & \textbf{.89/.91} & \textbf{67} & -           & -                 & -           & \textbf{18.5} & \textbf{.38}/.58         & 153           \\\hline
NLMs**                           & RNN cosine      & 12         & .48/.73          & 103         & 22          & 41/.70            & 116         & 69            & .28/.54         & 157           \\
                               & RNN w2v cosine  & 19         & .44/.70          & 111         & 19          & .44/.69           & 126         & 26            & \textbf{.38}/.66         & 111           \\
                               & RNN ranking     & 18         & .45/.67          & 128         & 24          & .43/.69           & 103         & 25            & .34/.66         & 102           \\
                               & RNN w2v ranking & 54         & .32/.56          & 155         & 33          & .36/.65           & 137         & 30            & .33/.69         & \textbf{77}   \\
                               & BoW cosine      & 22         & .44/.65          & 129         & 19          & .43/.69           & 103         & 50            & .34/.60         & 99            \\
                               & BoW w2v cosine  & 15         & .46/.71          & 124         & \textbf{14} & \textbf{.46/0.71} & 104         & 28            & .36/.66         & 99            \\
                               & BoW ranking     & 17         & .45/.68          & 115         & 22          & .42/.70           & \textbf{95} & 32            & .35/.69         & 101           \\
                               & BoW w2v ranking & 55         & .32/.56          & 155         & 36          & .35/.66           & 138         & 38            & .33/\textbf{.72}         & 85            \\             
\end{tabular}
Format: \textit{median rank \hspace{0.4cm} accuracy@10/100 \hspace{0.4cm} rank variance} 

*Low variance in mult models is due to consistently poor scores, so not highlighted.

** Hill et al \citeyear{hill2015learning} refer to their models as NLMs, this dissertation refers to them as definition models. 
\label{hill_results_1}
\end{table}

\begin{figure}[!htpb]
\centering
\includegraphics[width=1\linewidth]{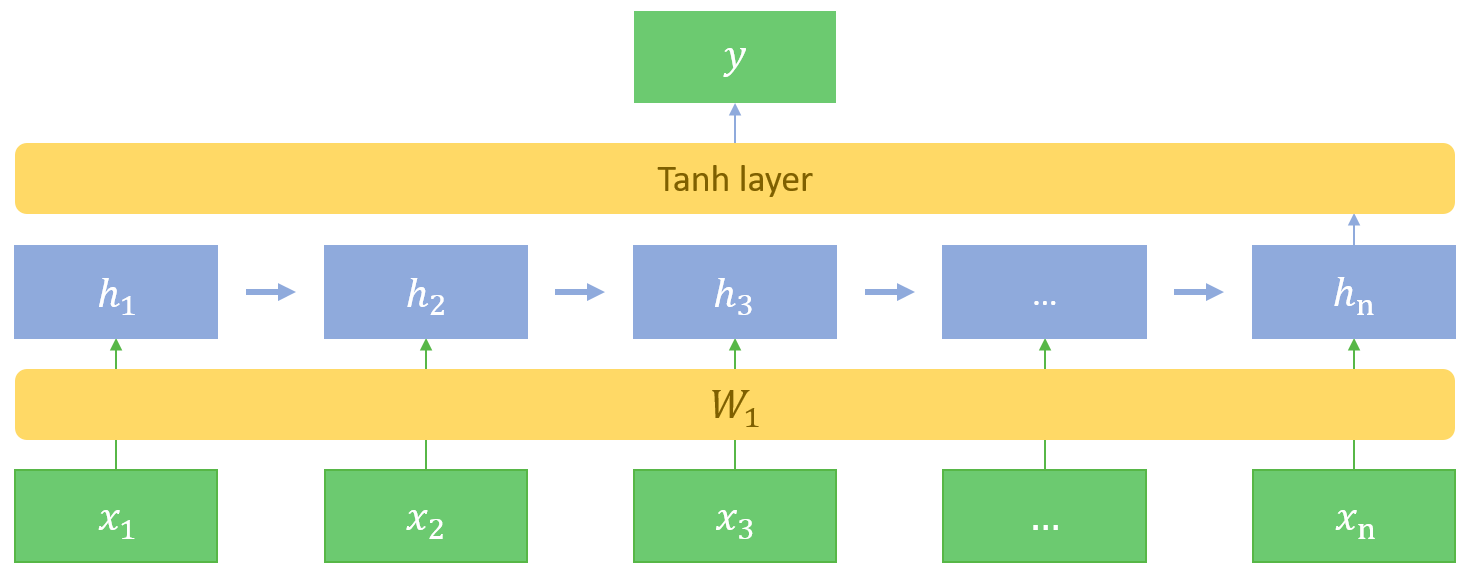}
\caption{Diagram of RNN based on Hill et al. (2015) implementation. $x_i$ … $x_n$ are the inputs, the integer represents the timestep $t$, $n$ is the length of the sequence, inputs are passed through the embedding layer $W$, $h_i$ … $h_n$ are the LSTM units which receive as input the embedding at time $t$ and the output from the previous LSTM hidden unit. The output of the final LSTM ($h_n$) is fed through a Tanh layer, projecting it into the dimensionality of the pretrained word embeddings. The output embedding is $y$, which is equal in dimensionality to the pretrained word embeddings.}
\label{hill_recurrent}
\end{figure}

The paper then explores two neural definition models with different ways of encoding the gloss, a recurrent neural model (RNN) and a neural bag-of-words model (BoW). For both models, to encode a gloss representation from learned embeddings, the integer representing the gloss word ID is used as an index for a linear projection weight matrix $W$ to find the learned word embedding $v_i$. Pretrained embeddings can be retrieved with a simple lookup; pretrained embeddings are used for all head words. For the RNN, the gloss representation is encoded by passing these word embeddings into an LSTM, which builds a representation of the entire sequence. At timestep $1$, just the first word is passed into the LSTM, then at all other timesteps, the LSTM cell takes as input the word at time $t$, and the previous LSTM hidden value \textit{$h_{t-1}$}. LSTM cells consist of three gates, input, forget, and output gates, each using two activation functions, a sigmoid to modulate the quantity of information that can pass through the gate and a tanh function to provide non-linearity. This means that there are three additional weight matrices to be learned during training, which are learned through backpropagation \cite[p.42]{DBLP:journals/corr/abs-1709-07809}. The final step is to pass the sequence-final output of the LSTM through a Tanh layer to add non-linearity and reduce dimensionality to match the pretrained head embedding. A diagram of the Hill et al. \citeyear{hill2015learning} RNN system can be seen in Figure \ref{hill_recurrent}.

\begin{figure}[!htpb]
\centering
\includegraphics[width=1\linewidth]{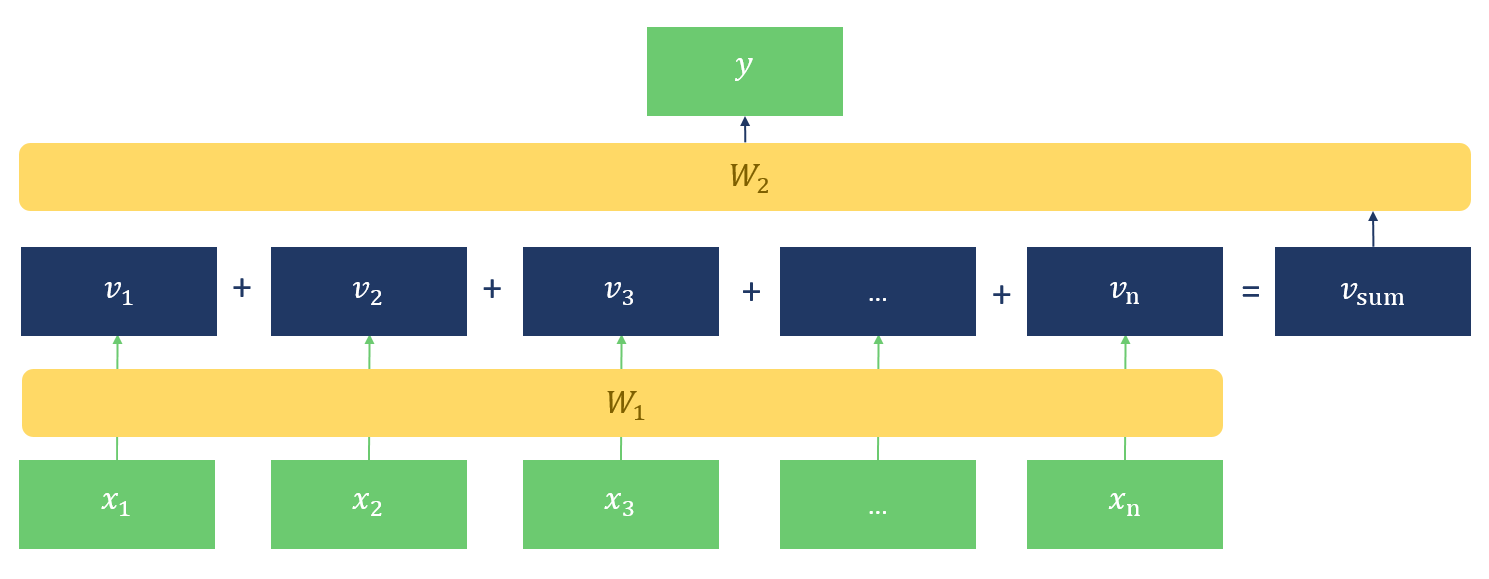}
\caption{Diagram of Neural BoW model based on Hill et al. (2015) implementation.
$x_i$ … $x_n$ are the inputs, the integer represents the timestep $t$, $n$ is the length of the sequence, the inputs are passed through the embedding matrix $W$ to produce word embeddings. $v_i$ … $v_n$ represent the word embeddings which are summed together to produce the output vector $v_{sum}$. This vector is passed through the tanh layer to change dimensionality. The output embedding is $y$, which is equal in dimensionality to the pretrained word embeddings.}
\label{hill_BoW}
\end{figure}

For the BoW model, glosses are encoded using a linear BoW. The model learns to map a definition with word embeddings $v_i...v_n$ to the sum of the projected embeddings. The sum of the projected embeddings $v_{sum}$ is then passed through a linear projection layer $W_2$ to reduce dimensionality as shown in Figure \ref{hill_BoW}.

\subsection{Training Objective}
Hill et al. \citeyear{hill2015learning} experiment with two different training objectives for the RNN and BoW models. The neural definition model $M$ is trained to map the gloss $s_c$ describing head word $c$ to an embedding space close to the pretrained embedding $v_c$ of $c$ \cite{hill2015learning}. The first training objective is simple cosine distance between the model output $y = M(s_c)$ and $v_c$. The second objective is rank loss, where the cosine distance of the model output is compared with both the correct word embedding $v_c$ and a random incorrect word in the model vocab $v_r$. The final equation is as follows:
\begin{equation}
$$
\centering $max(0, m + cos(M(s_c),v_c) - cos(M(s_c),v_r))$
$$
\label{correct_rank_loss}
\end{equation}
$m=0.1$ is a margin which ensures that in the event of $cos(M(s_c), v_c)$ and $cos(M(s_c), v_r)$ being equal the model is encouraged to learn.  
Note that the original equation in Hill et al. \citeyear{hill2015learning} is incorrect. Our notation (Equation \ref{correct_rank_loss}) is consistent with Huang et al. \citeyear{huang2012improving} and the later implementation described by this paper. 

The models trained with rank loss as the training objective perform best in a number of experiments, particularly for the concept descriptions task. Despite this, the training objective section in Hill et al. \citeyear{hill2015learning} is not very extensive. Furthermore, the experiments carried out in this dissertation do not find that rank loss outperforms cosine loss, which this dissertation suggests is because these models appear to take longer to converge.

Further implementation details for the models in Hill et al. \citeyear{hill2015learning} are listed below:
\begin{itemize}
\item Pretrained embedding length: 500
\item Learned embedding length: 256
\item Number of LSTM units: 512
\item Final hidden state mapped linearly to length 500, for comparison with target embedding
\item In BoW model: Projection matrix projects input embeddings (either learned=256 or pretrained=500) to length 500 for summation.
\item Models implemented in Theano
\item Trained with minibatch SGD on GPUs
\item Batch size of 16 and learning rate controlled by adadelta
\end{itemize}

\subsection{Tasks}
In accordance with the stated real-world applications of the models, Hill et al. \citeyear{hill2015learning} test all models on two separate tasks. Task one is a `reverse dictionary' task, head/gloss pairs are taken from dictionaries. Task two is a general-knowledge question task, where the model is tested on crossword clues in the place of dictionary definitions. For each task, the system output is compared with commercial systems designed to solve these tasks.

\subsubsection{Reverse Dictionary Task}

The first task on which the models are evaluated is a reverse dictionary task. The models are tested on the three definition test sets described in 2.1. For the purposes of evaluation, the system is compared with a commercial reverse dictionary system, \textit{OneLook.com}\footnote{https://www.onelook.com/reverse-dictionary.shtml}. The algorithm used by \textit{OneLook.com} searches 1061 dictionaries, meaning the effective database size is far larger than the training data for Hill et al.'s \citeyear{hill2015learning} model, and it requires internet access.

The results of these experiments are shown in Table \ref{hill_results_1}.

The three variables that differentiate the definition models are evaluated below: 
\begin{itemize}
\item \textbf{RNN vs. BoW:} The authors note that the RNN models do not achieve significantly better results than the BoW models. As is shown in the extensions section, the effectiveness of their RNN models may be compromised by the single forward direction of their LSTM encoder, and the sentence padding and truncating algorithms.
\item \textbf{Rank loss vs. cosine:} On the concept description task, for the models using learned embeddings for glosses, rank loss appears to outperform cosine loss in terms of median ranking and accuracy.
\item \textbf{Pretrained vs. learned embeddings:} Despite enjoying a smaller vocabulary, the W2V systems achieve better or at least comparable results on the test sets. The overall effect is not that significant, showing that the additional semantic information stored in the pretrained embeddings does not greatly outweigh the limited vocabulary that comes with using pretrained embeddings.
\end{itemize}

On both the `seen' and `concept description' test sets the \textit{OneLook} baseline achieves a lower (better) median rank than any of the models. The results on the seen test set are not surprising because these exact dictionary definitions may appear in the massive \textit{OneLook} database. The \textit{Onelook} results for the concept descriptions task are comparable to the definition models, despite having a lower median rank than any definition model, the accuracy @ 100 is surpassed by the BoW W2V ranking model. Qualitative analysis shows that the definition models return a more semantically consistent set of words, which share key properties of the description. In summary, the definition models perform admirably especially when qualitative results are compared with the more memory-expensive \textit{OneLook} system, but by most quantitative metrics, \textit{OneLook} is better.

\subsubsection{Crossword Answering Tasks}
Hill et al. \citeyear{hill2015learning} test the models on the crossword test set to see how well the models generalise to a novel task. It is noted that general knowledge crossword questions often mirror definitions in that they refer to fundamental properties of concepts, therefore this task is quite closely related to the definitions task. A unique advantage of crossword questions is that system outputs can be limited to those which are of the correct length (specified by the question), greatly reducing the output vocabulary, and making the correct word rank higher in the list. The advantages of this are discussed with reference to Figure \ref{vocab_distribution} in Section 4.

The definition models are evaluated against a commercial crossword-solving system \textit{Crossword Maestro}\footnote{http://www.crosswordmaestro.com/}. It is argued again that this system relies on database lookups at test time and some hand-engineered rules, something that the neural system does not require, potentially making the definition model developed by Hill et al. \citeyear{hill2015learning} more computationally efficient. All models were tested on the previously described \textit{Eddie James} crossword test set. The test set was split into three separate sets based on question length: 

\begin{itemize}
\item Long set (150 questions of length $>4$)
\item Short set (120 questions of length $<= 4$)
\item Single-word set (30 questions of length $1$)
\end{itemize}

Again two definition models are tested with and without pretrained W2V gloss embeddings, and with both rank loss and cosine loss objective functions, resulting in 8 total definition models. The results are in Table \ref{hill_results_2}.

\begin{table}[!htbp]
\caption{Crossword question results in Hill et al. (2015)}
\centering
\begin{tabular}{l|lll|lll|lll}
\toprule
                  & \multicolumn{3}{c|}{Long (150)}      & \multicolumn{3}{c|}{Short (120)}     & \multicolumn{3}{c}{Single-word (30)}        \\\midrule
OneLook           &    & .39 /            &             &    & .68 /            &             &            & .70 /            &             \\
Crossword Maestro &    & .27 /            &             &    & .43 /            &             &            & .73 /            &             \\\hline
W2V add           & 42 & .31/.63          & 92          & 11 & .50/.78          & 66          & \textbf{2} & \textbf{.79/.90} & 45          \\\hline
RNN cosine        & 15 & .43/.69          & 108         & 22 & .39/.67          & 117         & 72         & .31/.52          & 187         \\
RNN w2v cosine    & 4  & .61/.82          & 60          & \textbf{7}  & .56/.79          & 60          & 12         & .48/.72          & 116         \\
RNN ranking       & 6  & .58/\textbf{.84}         & \textbf{48} & 10 & .51/.73          & 57          & 12         & .48/.69          & 67          \\
RNN w2v ranking   & \textbf{3}  & \textbf{.62}/.80          & 61          & 8  & .57/.78          & 49          & 12         & .48/.69          & 114         \\
BoW cosine        & 4  & .60/.82          & 54          & \textbf{7}  & .56/.78          & 51          & 12         & .45/.72          & 137         \\
BoW w2v cosine    & 4  & .60/.83 & 56          & \textbf{7}  & .54/.80          & 48          & 3          & .59/.79          & 111         \\
BoW ranking       & 5  & \textbf{.62}/.87          & 50          & 8  & \textbf{.58/.83} & 37          & 8          & .55/.79          & \textbf{39} \\
BoW w2v ranking   & 5  & .60/.86          & \textbf{48} & 8  & .56/.83          & \textbf{35} & 4          & .55/.83          & 43         
\end{tabular}

Format: \textit{median rank accuracy@10/100 rank variance}. 
\label{hill_results_2}
\end{table}

The best median rank achieved by any model on the long question set was the RNN W2V ranking model. The correct answer ranked third on average, and was in the top-10 candidates 62\% of the time. As question length decreases, the commercial models' performance improves, so the largest improvement over the commercial models can be found on the long question set. The variables are analysed to more general effect below:

\begin{itemize}
\item \textbf{RNN vs BoW:} The recurrent models perform best on longer questions, but the BoW models are comparable even for the longer question set. For single-word questions the BoW models are much better, but their performance is surpassed by the simple W2V add model (and in this case there is only a single embedding so they are not summed). This is because single-word questions are often a case of finding synonyms eg. \textit{Guilt} ``Culpability" \cite{hill2015learning}, meaning direct comparison of word embeddings is the most effective method.
\item \textbf{Rank loss vs. cosine:} There appears to be little difference between the two loss metrics, although the best results on the long question set were attained using rank loss.
\item \textbf{Pretrained vs. learned embedings:} The best results are achieved with models using pretrained w2v embeddings but again the models are not dramatically different.
\end{itemize}

Qualitative analysis of the system shows again that it produces more consistent outputs than commercial models. A comparison shows that the commercial systems, while quite accurate, often output a word that has no semantic relation to the question, while the top 5-10 Hill et al. \citeyear{hill2015learning} system outputs were all somewhat related to the correct answer.

\subsection{Evaluating Hill et al.'s Work}
The impressive and consistent performance of the models on the crossword test and concept descriptions test shows that neural definition models can answer questions that are assumed to require ``general world knowledge'' and reasoning, something that neural networks are often believed to struggle with \cite{NIPS2017_7082}. For this dissertation, the extensions that will be applied to the system and explored in Section 4 are: 
\begin{itemize}
\item Including crossword question/answer data in the training set. This should adapt the system to better deal with the more obscure crossword questions that do not greatly resemble dictionary definitions, especially the shorter questions.

\item Using methods other than a single forward LSTM. Here we shall explore the use of a bidirectional LSTM and taking the average over LSTM states. The result should be a more balanced representation of the sentence that doesn't bias towards the final few tokens. This has also become somewhat standard in most neural QA tasks, as shown in the discussion of other related work.

\item Using sub-word units (defined below). Specifically, byte-pair encoding \cite{sennrich2015neural} will be applied to the glosses. This should help deal with out of vocabulary words and could result in a better gloss representation. 
\end{itemize}

\section{Related Work}
\subsection{Neural Language Modeling}
The models developed in this dissertation draw three main techniques from the field of natural language processing and neural machine translation: language models (outlined in the introduction), word embeddings, and sub-word level segmentation. 

\subsubsection{Word Embeddings}
Word embeddings are multi-dimensional vectors used to represent words. All langauge models including early N-gram models have to replace words with numbers in some form for processing. N-gram models simply replace words with their index in the vocabulary \cite{mikolov2013efficient}. In early use of word embeddings, the values in multi-dimensional word representations resulted from unsupervised clustering algorithms such as Brown clustering \cite{Turian:2010:WRS:1858681.1858721}, or more general distributional clustering \cite{Pereira:1993:DCE:981574.981598}. In neural language models of the type described in Bengio et al. \citeyear{bengio2003neural} and Koehn \citeyear{DBLP:journals/corr/abs-1709-07809}, word embeddings are stored in a matrix with dimension $V \times h$ where $V$ is the vocabulary size and $h$ is the size of each embedding. Each word in the vocabulary therefore has a unique $h$ dimensional embedding vector which can be retrieved by multiplying a one-hot vector representing the word with the embedding matrix. This is essentially a lookup of the matrix with the non-zero index of the one-hot vector \cite[p.34]{DBLP:journals/corr/abs-1709-07809}. This type of lookup is used for pretrained embeddings in the models in this dissertation. Of the models developed by Hill et al. \citeyear{hill2015learning}, most of them utilise both learned and pretrained word embeddings. For the pretrained embeddings, Hill et al.  \citeyear{hill2015learning} utilise embeddings created using the Google W2V software \cite{mikolov2013efficient} on over 8bn words of running text, which is then saved as a dictionary for lookup.

W2V software was developed by Mikolov et al. \citeyear{mikolov2013efficient}. Embeddings are learned using neural networks rather than any form of clustering. While there are two architectures, continuous bag-of-words (CBoW) and skip-gram, Hill et al \citeyear{hill2015learning} use the CBoW model, so the description here will be reserved to that. The model is based on a feedforward NNLM,
in which $N$ previous words are used as context words for the prediction of the current word. In the CBoW model, future context is also taken into account. There is a projection layer which is shared for all words. The model uses four past and four future context words, and because of the shared projection layer, the ordering of these context words is not considered. The model is trained by building a log-linear classifier, tasked with correctly classifying the current word based on these context words. During training, the projection layer is updated through backpropagation, thus creating an accurate embedding for each word.

For the models designed in this dissertation, GloVe embeddings \cite{pennington2014glove} are used. GloVe embeddings were chosen because pre-trained embeddings are easily sourced\footnote{https://nlp.stanford.edu/projects/glove/} and the 6B set shares the most similarities with the set described in Hill et al. \citeyear{hill2015learning}. The exact embeddings could not be replicated given time and resource constraints. GloVe embeddings work in a similar way to W2V, with adjustments made to the objective function that serve to improve performance on some similarity tasks as well as computational efficiency at training time \cite{pennington2014glove}.

\subsection{Neural Machine Translation}
Most research papers will acknowledge two main challenges in neural machine translation (NMT):
\begin{itemize}
\item Documents must be segmented into individual tokens in order to be represented numerically eg. using word embeddings.
\item Translation is an open vocabulary task, the system must be able to deal with any word in the language in question.
\end{itemize}
Here it is argued that definition models also have to consider these challenges. Therefore we draw upon at attempts to solve these problems in the field of NMT.

\subsubsection{Word-level Models}

The question of how to segment a document into individual tokens has been open since early attempts at probabilistic language modeling and statistical machine translation. The most simple approach is to segment on whitespace (known as a word-level model), but this can present issues for things like text normalisation and translation in instances when the individual words in a phrase should not be treated as individual words, such as phrase-based systems like that of Och et al. \citeyear{och1999improved}. For machine translation, word-level models can be suitable because they limit the size of the sequence to just the number of words in the sequence. Word-level NMT models, however, cannot model an unlimited vocabulary of words, which is especially problematic for morphologically-rich languages \cite{lee2016fully}. In these systems, words that are not in the system's limited vocabulary are usually assigned a special \texttt{UNK} symbol and translated at a later pipeline step. All of Hill et al's \citeyear{hill2015learning} models are word-level models, and some of the models implemented by this dissertation will also be word-level.

\subsubsection{Character-level Models}
Perhaps the most intuitive way of solving both of these problems is to use a character-level model. This solves the closed vocabulary problem by greatly limiting the number of possible tokens, and also eliminates the issue of segmentation. Character-level models however result in significantly longer sequences, making training such models incredibly costly; the character-level model of Luong and Manning \citeyear{luong2016achieving} took some three months to train. Further to this, a longer sequence means the RNN must pass information over a greater distance, which can lead to vanishing gradients \cite{sennrich2015neural}. A more efficient fully character-level NMT model was implemented by Lee et al. \citeyear{lee2016fully}. They believe that character-level models must be pursued, arguing that other existing methods of word segmentation for MT are sub-optimal because they are usually based on some rules that are not related to the task of translation \cite{lee2016fully}. They countered the long sequence problem with a multi-layered network of convolutional, pooling and highway layers. While their results are impressive, the model still takes some two weeks to train on a GPU, and therefore is not practical for this dissertation. Given, however, that the outputs of definition models are limited to one word (or a few words in certain crossword clues) we can explore character-level outputs, as in Section 6.

\subsubsection{Sub-word Units}
 
Due to the practical constraints of character-level models, many researchers find a compromise in sub-word units. These usually involve the application of some segmentation algorithm. Rule-based morphological segmentation algorithms such as Morfessor \cite{virpioja2013morfessor} have been used in statistical machine translation models for some years. Such segmentation algorithms have also seen some recent use in neural machine translation eg. \cite{machavcek2018morphological}. 

\begin{figure}[!htpb]
\includegraphics[width = 10cm]{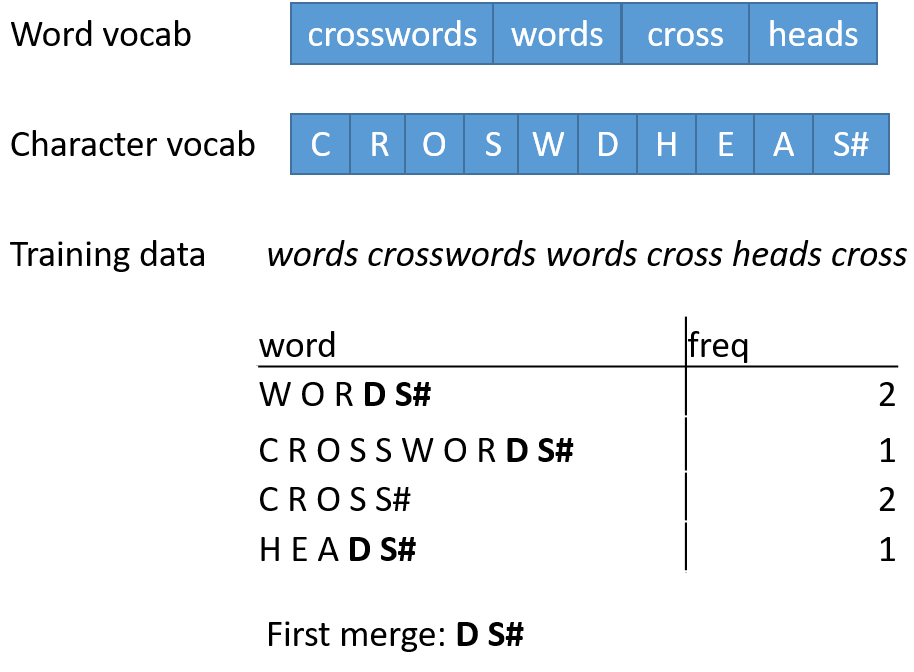}
\caption{First merge of BPE with toy training data}
\label{bpe_1}
\end{figure}

One of the most successful segmentation algorithms for sub-word NMT is byte-pair encoding (BPE) \cite{sennrich2015neural}. The first step in BPE as applied to the task of word segmentation is to create a vocabulary of all characters and symbols in the training corpus, as shown in Figure \ref{bpe_1}. Note that the BPE training corpus does not have to be the training data used for experiments.

\begin{figure}[!htpb]
\includegraphics[width = 10cm]{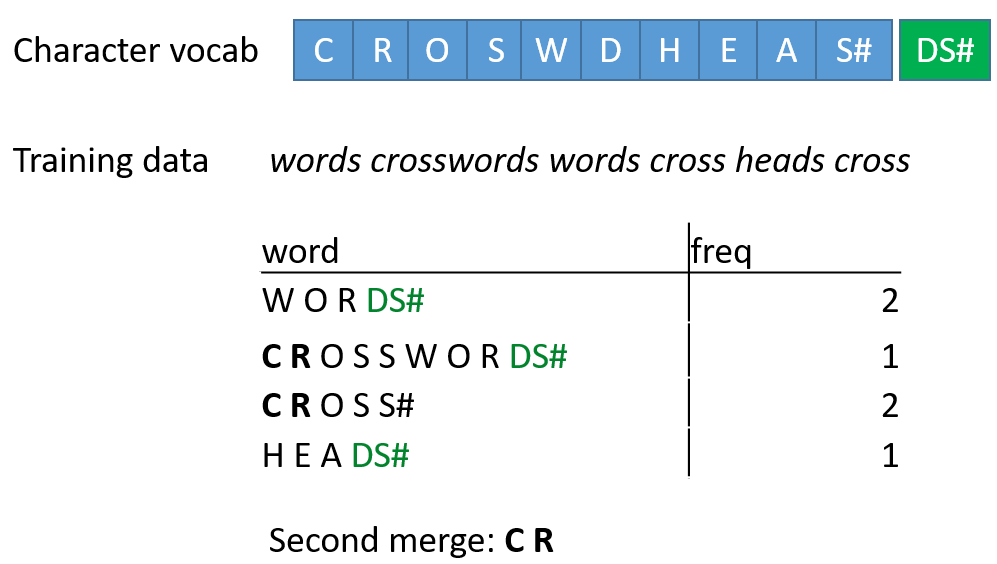}
\caption{Second merge of BPE with toy training data}
\label{bpe_2}
\end{figure}

\begin{figure}[!htpb]
\includegraphics[width = 15cm]{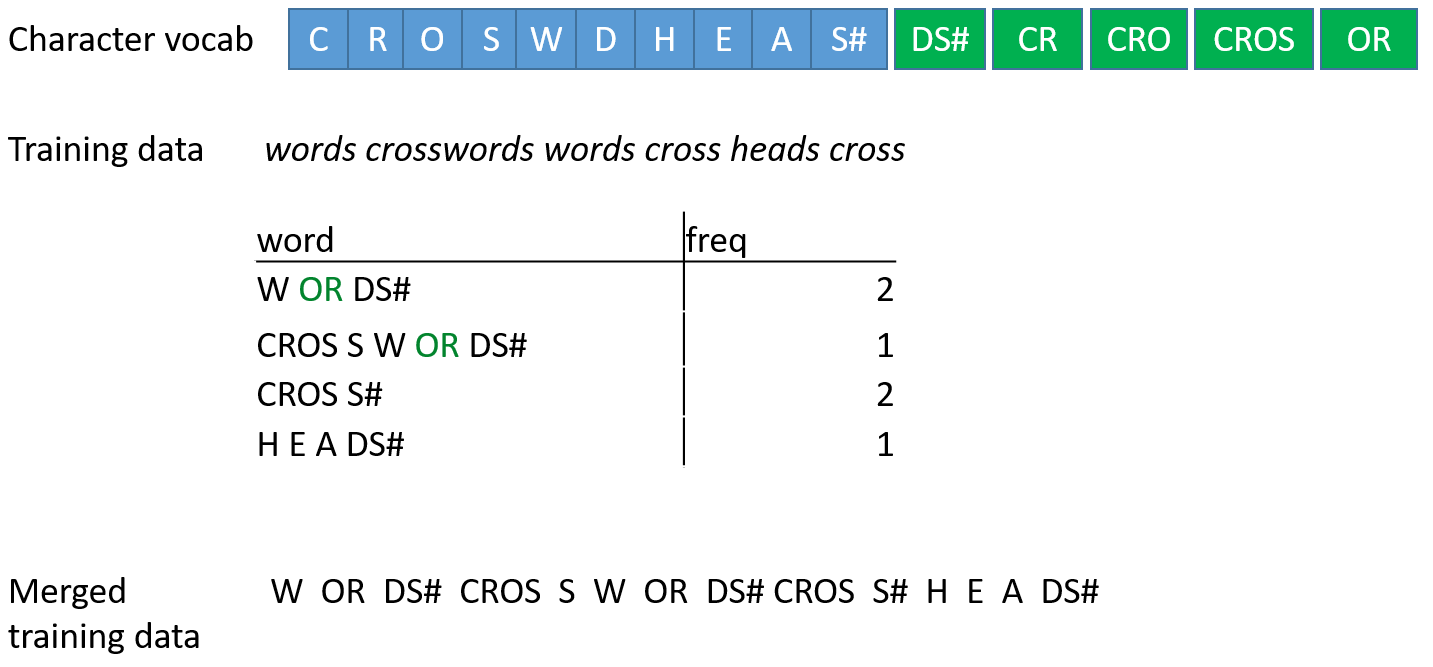}
\caption{Toy training data after five total merge operations}
\label{bpe_3}
\end{figure}

In Figure \ref{bpe_1} the \# symbol represents a word-final character. On each pass, the algorithm merges the two most frequent adjacent symbols (symbol pair) observed in the training data into a single vocabulary entry. In the example, the first merge produces the symbol DS\# which is added to the dictionary. The algorithm is recursive and stops after a set number of merges (stopping criteria). After each symbol is added to the dictionary, the data is re-segmented and counts are taken for the new vocabulary. Figure \ref{bpe_2} shows the data after a second merge, and Figure \ref{bpe_3} shows the data after five total merge steps. Note that when there is an equal number of occurrences of two symbol pairs, the left symbol earliest in the alphabet is preferred.

It is argued that BPE lends itself to NMT because it is less conservative than some other morphological segmentation algorithms commonly used in phrase-based statistical machine translation \cite{sennrich2015neural}. Sennrich et al. use BPE to surpass the back-off dictionary baseline results for the WMT 15 translation task, proving its effectiveness. Machavcek \citeyear{machavcek2018morphological} makes some suggestions to improve BPE for their task (Czech segmentation), which are not explored here, but could be interesting to explore in future work on definition models. In this dissertation we apply BPE to glosses to allow a wider input vocabulary and better learned embeddings.

\subsection{Neural Question Answering}
Definition models do not take questions as input, rather definitions (or glosses), and they seek only to output single words. Question answering systems (as opposed to definition models) take full questions as input and output a full sentence in natural language as the answer. This type of system was developed by Yin et al. \citeyear{yin2015neural}. This model is much larger than that of Hill et al. \citeyear{hill2015learning} in every respect, it is trained on considerably more data and is more complex. The main differences are that Yin et al. \citeyear{yin2015neural}:

\begin{itemize}
\item Enlist the use of a knowledge base of facts for retrieval at test time.
\item Train on a much larger corpus (235m QA pairs).
\item Implement a decoder capable of outputting full natural language sentences.
\end{itemize}

The system ``interprets'' the sentence using a bidirectional RNN, ``enquires'' the knowledge base by scoring the question and ranking candidate answers, then ``answers'' the question in a natural language sentence with the use of a decoder \cite{yin2015neural}. The decoder is highly effective at producing intelligible outputs as it is able to switch between generating a common word and outputting a term retrieved from the knowledge base with a certain probability \cite{yin2015neural}.

The system developed by \cite{yin2015neural} intelligently combines three separate models to create an end-to-end question answering system. The scale of their system is far beyond the scope of this dissertation and therefore we cannot adopt much of their implementation. The use of a bidirectional LSTM as a gloss encoder will however be explored.

\begin{figure}[!htpb]
\centering
\includegraphics[width=1\linewidth]{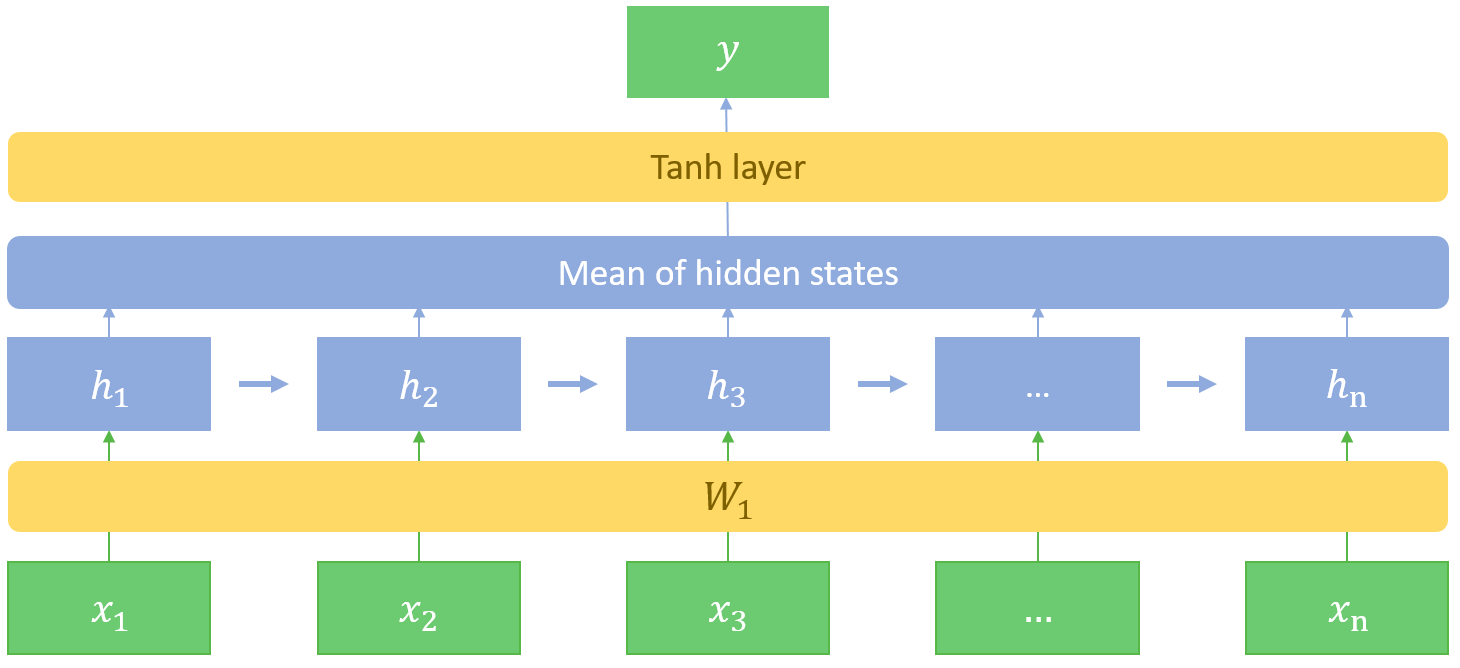}
\caption{Diagram of RNN model using mean of LSTM states.
The diagram is identical to Figure \ref{hill_recurrent}, except the outputs of the LSTM states at each timestep ($h_i$ … $h_n$) are averaged to form a mean vector before being passed through the Tanh layer.}
\label{mean_diagram}
\end{figure}

Bachrach et al. \citeyear{bachrach2017attention} experiment with various LSTM architectures in their neural QA system. Instead of taking the final LSTM state or concatenation of the final states of both LSTMs (in a bidirectional system), they propose to take the coordinate-wise mean of the LSTM hidden states at each timestep. This can be visualised in Figure \ref{mean_diagram}.

The goal of both the bidirectional LSTM and the average of LSTM states is to reduce the effects of vanishing gradients and loss of information which came early in the sequence. The single LSTM model as implemented by Hill et al. \citeyear{hill2015learning} would likely ``forget'' information that came at the beginning of the sequence as the information is overwritten by new information passing through the input gate. This dissertation hypothesises that taking the average across the hidden states of a single LSTM may serve to slightly bias towards information at the beginning of the sequence, as this information has a chance to be passed all the way through the LSTM via the cell state, meaning it could have an influence across LSTM hidden states for the entire sequence. Given that we have identified the use of a single forward LSTM as a potential area to improve the Hill et al. \citeyear{hill2015learning} model, we experiment with both a bidirectional LSTM and taking the average across the LSTM states.

\section{The Crossword Answering Model}
\subsection{Data}
The initial model developed for this task is as faithful to the definition model developed in Hill et al. \citeyear{hill2015learning} as possible. The original implementation is no longer available, therefore this model has been developed using a version of the code which was adapted for a Cambridge University masters course\footnote{https://github.com/fh295/Cambridge\_course.git}. This implementation is slightly different, and there are also some notable differences to the datasets. The training data is consistent with the paper, $\sim850,000$ word definition pairs, with $\sim90,000$ unique head words. An additional test set of crossword questions has also been gathered and is combined with this definition training set for some experiments. The test set is smaller than that described in the paper, consisting of only 200 concept descriptions, compared with the four separate sets described in Hill et al. \citeyear{hill2015learning} (listed in 2.1).

Results will be considered under the assumption that the training set provided is the same as the one described in Hill et al \citeyear{hill2015learning}. Regarding pretrained word embeddings, some are provided in the Cambridge course, but they do not cover the set of head words in the training data, meaning the model can not reliably learn representations even for words in the training data. The embeddings described in the paper are trained using Word2Vec (W2V) software \cite{mikolov2013efficient} on 8bn words of running text, so to best replicate this, GloVe \cite{pennington2014glove} 6B embeddings\footnote{https://nlp.stanford.edu/projects/glove/} are used in place of the incomplete embeddings provided. A discussion of these different types of word embeddings is found in Section 2.1.

\subsection{Implementation}
While the original model is implemented in Theano \cite{bergstra2010theano}, this model is implemented using Tensorflow \cite{abadi2016tensorflow}. An RNN is used to create a representation of the input sequence. For an input word sequence of length $n$ ($w_1, w_2 ... w_n$), at each timestep $t$, the RNN unit takes as input both $w_t$ and $h_{t-1}$ where h is the hidden state of the RNN. 

Because the code was designed for an assignment, it was not fully implemented. Completion of the cosine similarity evaluation function was required to attain results for the median rank on the test set. The model was also extended with an option of rank loss as an objective function as defined by Equation \ref{correct_rank_loss}.

\begin{figure}[!htpb]
\centering
\includegraphics[width=0.75\linewidth]{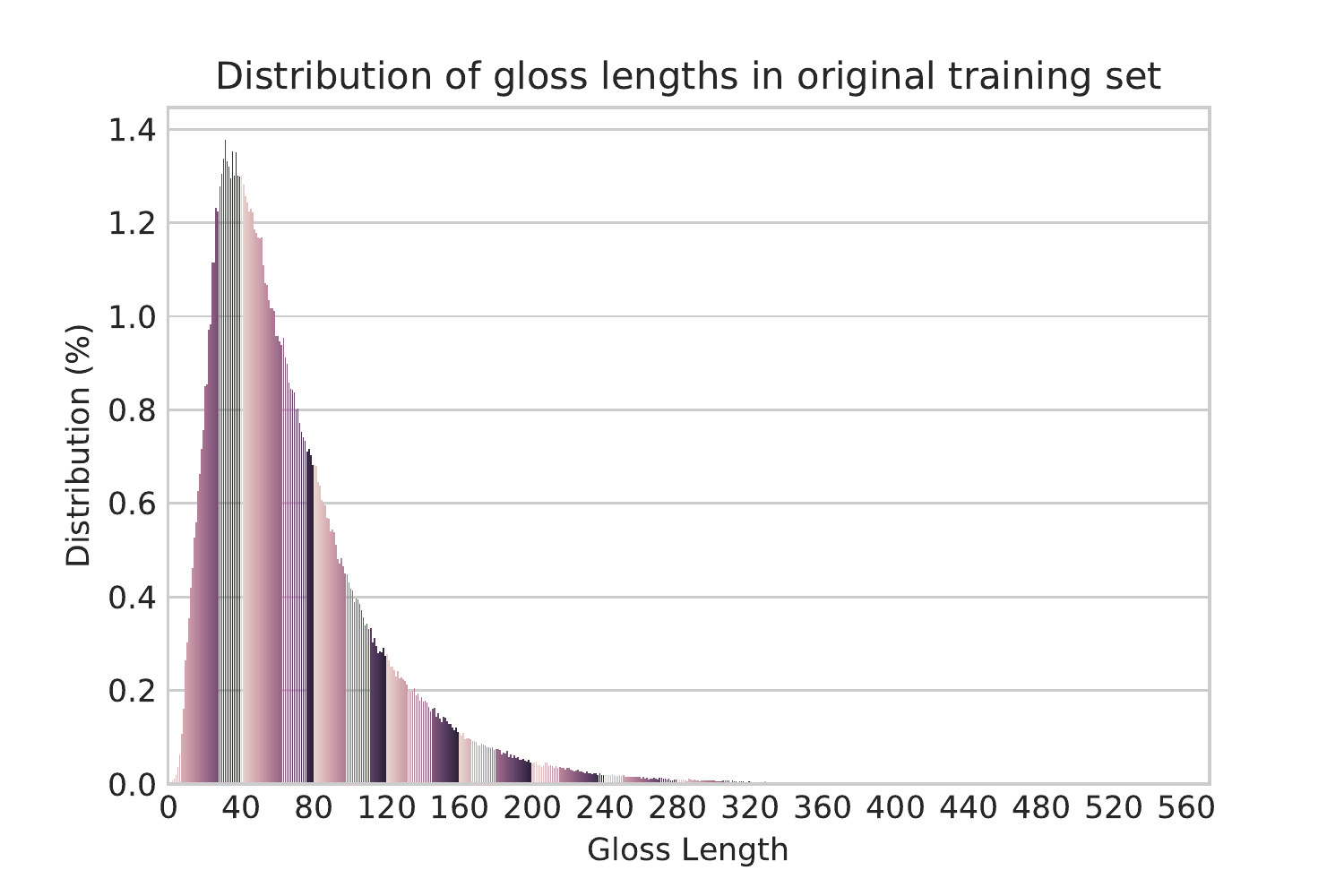}
\caption{Gloss length distribution in original training file}.
\label{s_d_lengths}
\end{figure}

\begin{figure}[!htpb]
\centering
\includegraphics[width=0.75\linewidth]{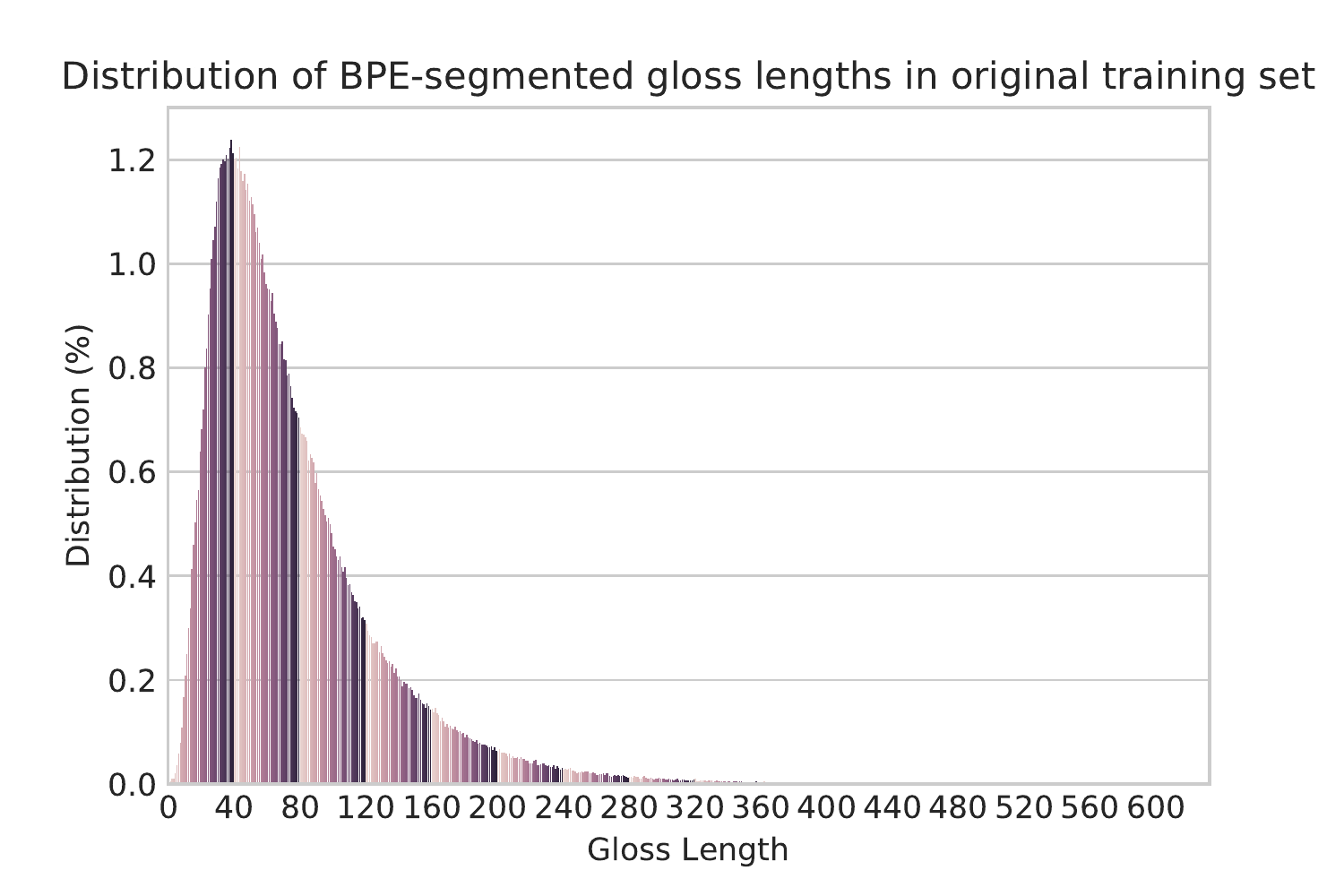}
\caption{Gloss length distribution in BPE-segmented original training file}.
\label{s_b_lengths}
\end{figure}

\begin{figure}[!htpb]
\centering
\includegraphics[width=0.75\linewidth]{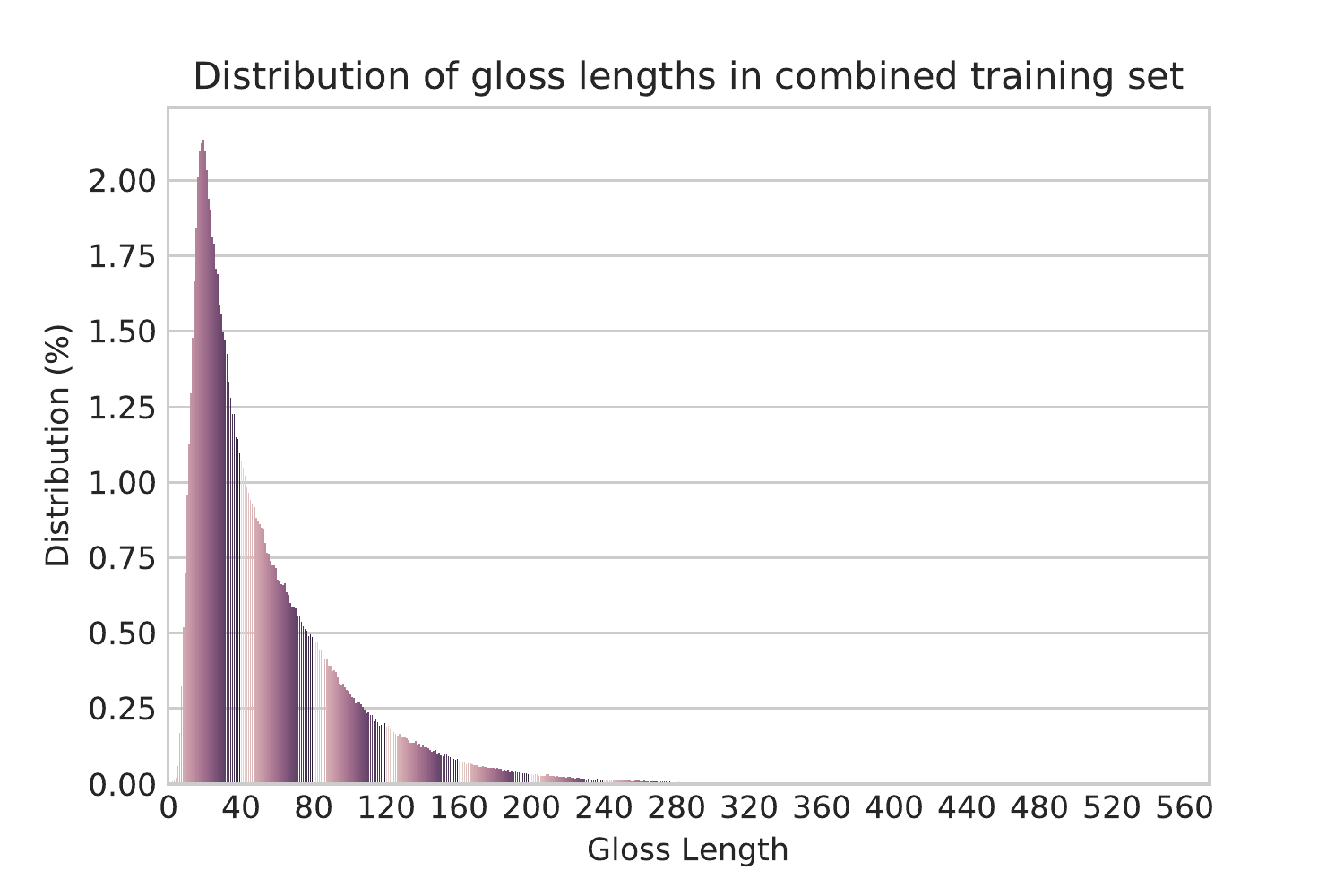}
\caption{Gloss length distribution in combined training file (includes crossword data)}.
\label{l_d_lengths}
\end{figure}

\begin{figure}[!htpb]
\centering
\includegraphics[width=0.75\linewidth]{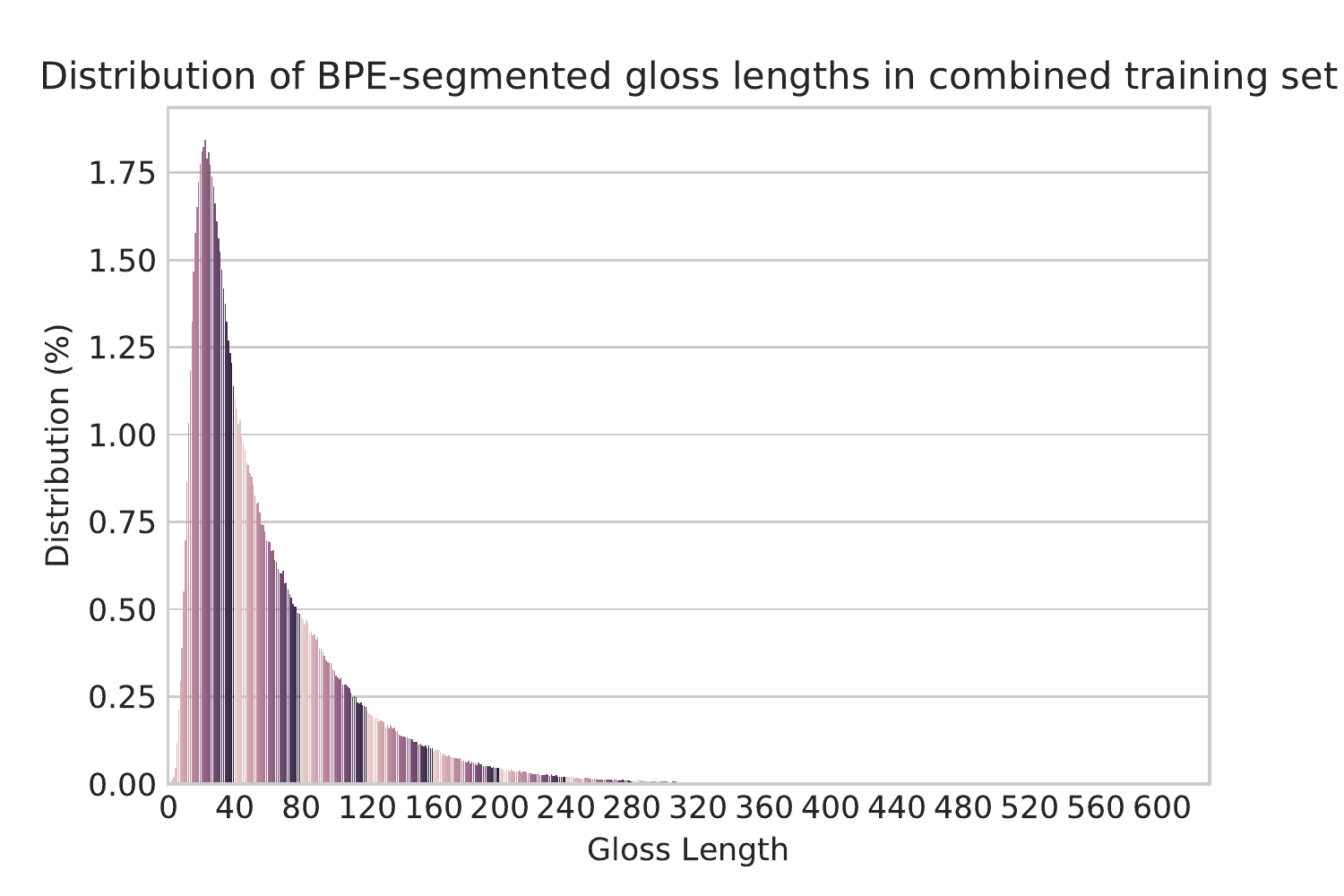}
\caption{Gloss length distribution in BPE-segmented combined training file (includes crossword data)}.
\label{l_b_lengths}
\end{figure}

Note that in the Hill et al. \citeyear{hill2015learning} model, sequences are either padded or truncated to be equal length (20). The distribution of gloss lengths in the different training datasets is shown in Figures \ref{s_d_lengths}, \ref{s_b_lengths}, \ref{l_d_lengths} and \ref{l_b_lengths}. The original training set, used by Hill et al \citeyear{hill2015learning} contains many glosses longer than 20 words, with an average gloss length of around 40, as shown in \ref{s_d_lengths}. This is certainly grounds for including longer sequences and not truncating to length 20.

As a modification, the experiments carried out here pad glosses in a less uniform way, to preserve as much information as possible in the glosses. This is not a problem as RNNs can take variable length inputs and the output representation will always be a vector of the same dimensionality. The process of padding/truncating no longer takes place as a preprocessing step, instead it is implemented dynamically during training, this does lead to slightly increased CPU load. Glosses are only padded up to the length of the longest sequence in their minibatch. To facilitate sequences of uniform length, the training data was split into large batches of 4096, then sorted from shortest to longest within these batches. Splitting into batches like this was seen as preferable to sorting the entire dataset, which would mean the model trained on gradually longer sequences throughout each epoch, which could cause a large increase in loss at the beginning of each epoch. The large batch size is chosen to be divisible by the minibatch size used in all experiments (16). The maximum padding size is set at 150. This change allows for no padding to be applied to the test examples.

Most of the experiments run here use learned input embeddings and pretrained head embeddings, while it was found in Hill et al. \citeyear{hill2015learning} that models using pretrained gloss embeddings perform best on some test sets, the difference is not hugely significant, and the BPE units learned on the training data are not represented in the pretrained embeddings, meaning the rate of unknown words would be much higher on BPE models. Input embeddings are be learned from the training data using the embedding lookup function. The maximum number of embeddings to be learned is fixed to 100,000 for the word-level experiments, meaning the 100,000 highest frequency words in the training data are used, and an \texttt{UNK} symbol is added to the vocabulary for out of vocabulary (OOV) words. For BPE experiments, the vocabulary size depends on which of the two training sets is used. In both cases this is $<100,000$ and includes every token in the final segmented training set. 

\subsection{Models}

In order to give the experiments a tighter focus, and because this dissertation believes there is more scope to improve the RNN models, the NBoW implementation is not explored in these experiments; all models will use RNNs. The basic implementation, therefore, looks exactly like the model in Figure \ref{hill_recurrent}. The other models are differentiated by loss function (cosine/rank), segmentation (word-level/BPE), and training data (definitions/full), resulting in 8 total models shown in Tables \ref{overall_results} and \ref{crossword_results}. All of these models are trained using LSTM average after this was chosen as the best encoder type following comparison of single RNN, bidirectional RNN and LSTM average (shown in \ref{bidirectional_comparison} and discussed in 5.4).

\begin{figure}[!htpb]
\centering
\includegraphics[width=1\linewidth]{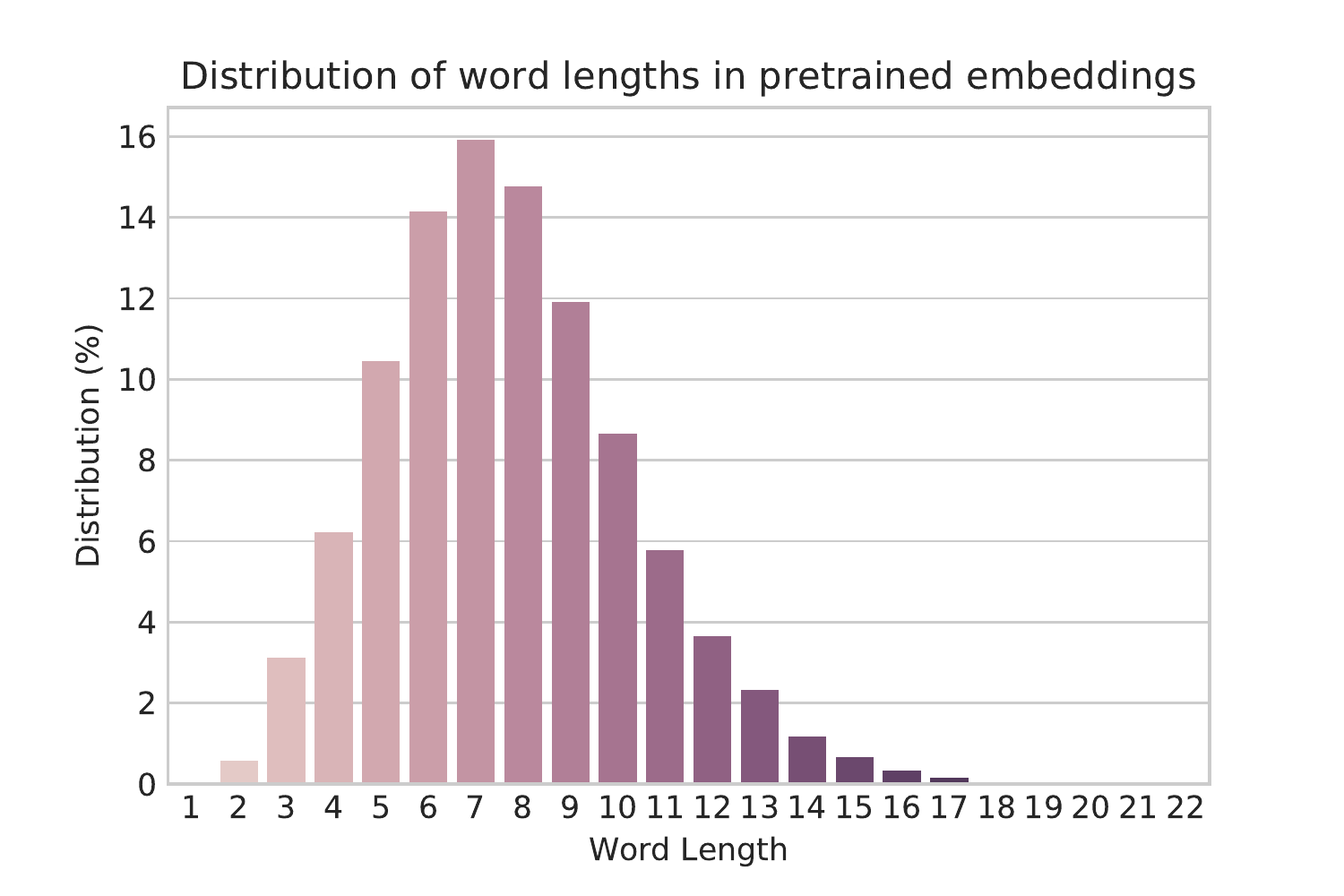}
\caption{Word length distribution in embeddings file}.
\label{vocab_distribution}
\end{figure}

For reporting of all results on the crossword test sets, the evaluation function was edited to reduce the output vocabulary to only words of the correct length. To gain insight into how much this reduces the search space, the distribution of words in the embeddings file (which also acts as the output vocabulary) is shown in Figure \ref{vocab_distribution}. This figure shows that the most common word length is 7, but even if the correct head word is length 7 the search space is reduced to under 16\% of the entire output vocab.

\section{Experiments}
\subsection{Hardware}
Unless otherwise stated, all models trained for this paper have been trained remotely on the Edinburgh Compute and Data Facility GPU clusters known as ``Eddie''\footnote{https://www.ed.ac.uk/information-services/research-support/research-computing/ecdf}. The GPUs available are Tesla K80 and Nvidia TitanX. Training was carried out on one of these, subject to availability.

\subsection{Reverse Dictionary Tasks}
In the interest of remaining as faithful to Hill et al. \citeyear{hill2015learning} as possible, the model was first optimised on the reverse dictionary task. The training data is the full $\approx850,000$ definitions set as described in Hill et al. \citeyear{hill2015learning}. The test set is the user-generated ``concept descriptions'' test set defined in 3.1. The other test sets in Hill et al. \citeyear{hill2015learning} were not included in the repository.

The batch size was chosen according to Hill et al. \citeyear{hill2015learning}, other parameters were optimised through grid search:
\begin{itemize}
\item Learned embedding size: 500
\item Learning rate, optimiser: 0.0001, Adam
\item Epochs: 10
\item Batch size: 16
\end{itemize}

A comparison of an attempt to reproduce the exact model of Hill et al \citeyear{hill2015learning} on the definitions training set can be found in the Table \ref{hill_jp_comparison}. Both models were trained on the definitions training set at word-level with cosine loss as an objective function, and tested on the concept descriptions test set

\begin{table}[!htpb]
\caption{Comparison of gloss encoder architectures}
\centering
\begin{tabular}{ll}
\toprule
Model     & Median Rank (concept descriptions) \\\midrule
Hill et al RNN cosine    & 69.0             \\
Our RNN cosine       & 106.5               \\
\end{tabular}
\label{hill_jp_comparison}
\end{table}

A comparison with the results in Table \ref{hill_results_1} shows that their RNN cosine model achieved a median rank of 69 on the concept descriptions task, while our model attained a median rank of 106.5 on the same test set. The top 10 and top 100 performance can not be reported for this model, but it is expected that it would also have performed worse. This discrepancy could possibly stem from one of the following:
\begin{itemize}
\item The use of GloVe embeddings rather than W2V.
\item A difference in hyperparameters such as the learning rate optimiser.
\item Some other difference in the code or datasets that would be impossible to uncover without access to the original implementation .
\end{itemize}

With these points in mind, it is noted that we do not have access to the original implementation, and cannot further optimise the hyperparameters due to resource constraints. Despite this discrepancy, the results are quite close, and this dissertation posits that any improvement made to these models would also have positive effects on Hill et al.'s models.

\subsection{Crossword Answering Tasks}
As explained, the focus of this work is to improve the model specifically for the crossword answering task. In the Cambridge assignment repository there was no crossword question test set, and the website\footnote{eddiejames.co.uk, now unavailable} cited in the paper as the resource for this test set has since been taken down. New data was therefore collected for this task. Note that in accordance with Hill et al. \citeyear{hill2015learning} all crossword clues are ``quick" or ``general knowledge" clues, as opposed to ``cryptic" clues. These were chosen due to their similarity with normal dictionary definitions. Crossword questions and answers were gathered from two sources:
\begin{itemize}
\item[1.] A Github repository containing $\sim600,000$ clues and answers taken from the New York Times crosswords\footnote{https://github.com/donohoe/nyt-crossword, accessed 02/18}.
\item[2.] A dataset of $\sim120,000$ clues and answers gathered from The Guardian website\footnote{https://www.theguardian.com/crosswords/series/quick} using beautifulsoup and requests. 
\end{itemize}

The datasets are cleaned by removing all exact duplicate clue/answer pairs, removing any answers that contain multiple words, and split into separate ``long'' and ``short'' question sets in accordance with Hill et al. \citeyear{hill2015learning}. Long questions are those consisting of $>4$ words, and short questions are $<=4$ words. These word counts were taken after removing special characters but before any sub-word segmentation was applied. Note, single-word clues are included in the ``short'' category for ease of reporting. The contents of the final datasets are summarised in table \ref{crossword_data}. 

\begin{table}[!htbp]
\centering
\caption{Summary of crossword data gathered for experiments}
\begin{tabular}{lll}
\toprule
Source   & Dataset                      & Final count \\\midrule
Guardian & long (\textgreater{}4 words) & 14,421      \\
         & short (\textless{}=4 words)  & 59,127      \\\midrule
NYT      & long                         & 48,933      \\
         & short                        & 263,098    
\end{tabular}
\label{crossword_data}
\end{table}

Unfortunately removing all multiple-word answers significantly reduced the size of these datasets, and the cleaning process overall resulted in around a 50\% reduction in size of both sets. Despite this, the final number of usable crossword questions gathered for this paper is 385,579 which certainly should be enough to have a noticeable effect on training.

The test set was extended to include 100 long questions and 100 short questions extracted from each dataset, totaling 400 additional test examples. The overall test results for the all final models are shown in Table \ref{overall_results}, and median rank result split by crossword test set is shown in Table \ref{crossword_results}. Note that all of these models were trained taking the average of LSTM states (see Section 5.4.2), which proves to be an improvement over the basic implementation (a single forward LSTM). For an explanation of applying BPE see Section 5.4.1.

\begin{table}[!htbp]
\caption{Overall results}
\begin{tabular}{l|ll|ll}
\toprule
                             & \multicolumn{2}{c|}{Concept descriptions}      & \multicolumn{2}{c}{Crossword overall}            \\
Model (dataset)              & Median rank   &\% in top 10/100    & Median rank    &\% in top 10/100     \\\midrule
RNN cosine (definitions)     & 72.5         & 23.5/52.5          & \textbf{304.5}          & \textbf{21.5}/\textbf{34.25}           \\
RNN rank (definitions)       & 243.0         & 12.5/40.0          & 743.0         & 6.5/24.0          \\
RNN BPE cosine (definitions) & \textbf{61.0} & \textbf{29.5}/\textbf{55.5} & 474.0 & 14.75/31.75 \\
RNN BPE rank (definitions)   & 191.5         & 10.5/38.0          & 750.5          & 8.0/24.0           \\\midrule
RNN cosine (full)            & 55.0          & 31.0/\textbf{59.0}         & \textbf{50.5}           & \textbf{35.75}/\textbf{56.25}           \\
RNN rank (full)              & 90.0         & 17.0/52.5         & 98.0           & 24.25/50.5         \\
RNN BPE cosine (full)        & \textbf{53.0}          & \textbf{34.0}/57.0          & 65.5           & 30.0/54.75         \\
RNN BPE rank (full)          & 70.5          & 21.5/54.5         & 82.5           & 23.5/52.5
\end{tabular}
\label{overall_results}
\end{table}

Firstly it is noted that the cosine model performs considerably better than the rank loss model, contrary to the results found by Hill et al. \citeyear{hill2015learning}. This is a surprising result, but it could be suggested that this model was taking longer to converge due to the more ambiguous training objective, and should have been trained for more than 10 epochs, as shown in Figure \ref{convergence}.

\begin{table}[!htbp]
\caption{Crossword results}
\begin{tabular}{l|ll|llll}
\toprule
                             & Guardian long  & Guardian short & NYT long        & NYT short      \\
Model (dataset)              & Median rank    & Median rank    & Median rank     & Median rank    \\\midrule
RNN cosine (definitions)     & \textbf{121.5}          & \textbf{140.5}          & \textbf{763.0}          & \textbf{531.5}          \\
RNN rank (definitions)       & 260.5          & 349.5         & 1069.0          & 1169.5         \\
RNN BPE cosine (definitions) & 178.0 & 211.0 & 1280.5 & 649.0 \\
RNN BPE rank (definitions)   & 247.5          & 669.5          & 1601.5          & 1149.0         \\\midrule
RNN cosine (full)            &\textbf{14.5}&\textbf{6.5}&270.0&\textbf{109.5}\\
RNN rank (full)              &44.5&20.5&389.5&233.0\\
RNN BPE cosine (full)        &19.0&11.0&414.0&110.5\\
RNN BPE rank (full)          &24.5&40.5&\textbf{247.5}&129.5
\end{tabular}
\label{crossword_results}
\end{table}

For models trained on only the definitions training set, the results on all crossword test sets were worse than the concept description task, this is probably because the question style is quite different to the definitions included in the training data. The New York Times dataset is by far the most cryptic, hence the relatively poor performance across all models for this dataset. Here is a comparison of the three question styles, as they appear cleaned and lowercased in the training data: 

\begin{itemize}
\item Definitions: \textit{restored} ``to bring back to its former state to bring back from a state of ruin decay disease or the like to repair to renew to recover''
\item Guardian: \textit{restore} ``give back return to former state'' 
\item NYT: \textit{restored} ``made as good as new''
\end{itemize}

In order to try to improve the results, the 385,579 crossword questions were added to the training set to create a final training dataset of over 1.2m head-gloss pairs. The results of the two word-level models on the full dataset are presented in Tables \ref{overall_results} and \ref{crossword_results}. Table \ref{crossword_results} shows a significant improvement for models trained on the full dataset on all of the crossword test sets which is exactly what was expected, because the additional training data should much more closely resemble these test examples. It is also interesting to note that the performance on the concept descriptions test set has improved too. This could possibly be explained by the non-professional human written nature of the concept descriptions test set, which could arguably be more similar to crossword questions than the professionally-written dictionary definitions training set. A further possible explanation is simply that including more relatively similar training data, even if it does not very closely resemble the test data, has a positive effect on the network's ability to compose glosses into head words.

\subsection{Model Extensions}
This section considers ways to extend the implementation of the model in order to improve performance on both definition and crossword tasks. Inspiration for these extensions has been drawn from recent research into neural question answering systems and neural sequence-to-sequence models such as translation systems, all of which were discussed in the related work section.

\subsubsection{Byte pair encoding}
This extension tested the application of BPE (as described in 3.2.3) to the training and testing process. The BPE unit vocabulary was first learned on the training data with the number of merges set to 10,000. This is slightly lower than the recommended number of merges because the training data is relatively small. An example of how the merging process works is given in 3.2.3 Despite the relatively small number of merges, many of the definitions still remained completely intact as whole words. This learned BPE vocabulary was applied to the glosses in both the training and test sets. The head words remained whole words, and pretrained embeddings were used for these. The set of all tokens in the definitions training set was reduced through this process from $100,000$ to $\sim75,000$ words and sub-word units, meaning the system should not have to deal with any unknown words or sub-word units in the glosses.

\begin{figure}[!htpb]
\centering
\includegraphics[width=1\linewidth]{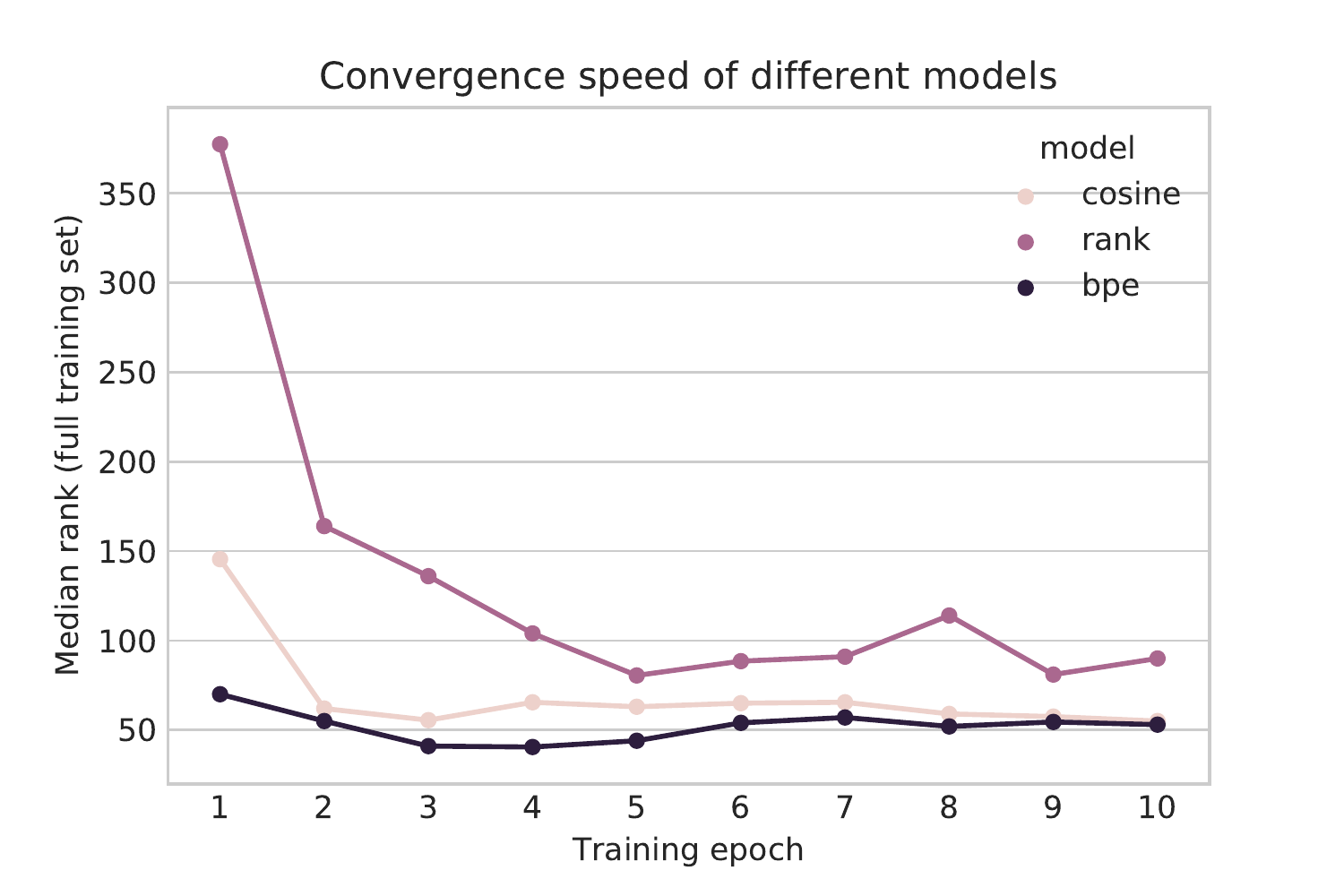}
\caption{Line graph showing dev set performance over time of three models. The BPE model converges fastest, with a lowest median rank at epoch 3, at which point it begins to overfit the training data. The cosine model's lowest median rank is at epoch 10, and the rank loss model doesnt appear to converge.}
\label{convergence}
\end{figure}

An interesting finding from this is that the models trained using BPE converged at an optimum significantly faster than the word-level models, as shown in Figure \ref{convergence}. 

This graph shows training on the full dataset. Convergence is clearly fastest for the BPE model. This was also true for models trained on the smaller dataset; the reduction in loss per batch, and improvement on test set performance per epoch is higher on all BPE models. 

This had a positive effect on the concept descriptions task, as shown in Table \ref{overall_results}, where the best results achieved for models on both training sets were BPE models. Note that all of these models were trained for 10 epochs, and this led to some overtraining for the BPE models. This may explain the worse performance for the BPE models on the crossword tasks (see Table \ref{crossword_results}). This highlights the fact that not only were the BPE models better on the definitions set, they also trained faster, and should have only trained for around 3 epochs. Given more time, it would also be an interesting experiment to try different numbers of merges to see if the performance could be improved further.

\subsubsection{Taking average of LSTM outputs}
It was explained earlier that the preferred type of RNN for language modelling tasks such as this is an LSTM. The reason for this is that LSTMs tend to limit the effects of memory loss through vanishing gradients by adjusting how much of the existing representation is replaced by new information. That being said, LSTMs can still bias towards information that comes at the end of the sequence, so the final LSTM state may not be a faithful representation of the entire sequence. Hill et al. \citeyear{hill2015learning} do not experiment with other ways of using RNNs to combine the gloss sequence.

Here, instead of using the final LSTM state to represent the sequence, the mean average is taken across LSTM states at each timestep. The model can be visualised in Figure \ref{mean_diagram}. This technique aims to give more equal weighting to all words in the sequence. This is similar to the use of LSTMs in the neural QA model of Bachrach et al. \citeyear{bachrach2017attention}, as described in 3.3.

This proved to have a positive effect, as shown in Table \ref{bidirectional_comparison}. The models trained with LSTM average perform better than the single forward LSTM. The LSTM average also appeared to perform quite well on shorter questions, which is why it was selected for the models in Tables \ref{overall_results} and \ref{crossword_results}.

\subsubsection{Using a bidirectional LSTM}
\begin{figure}[!htpb]
\centering
\includegraphics[width=1\linewidth]{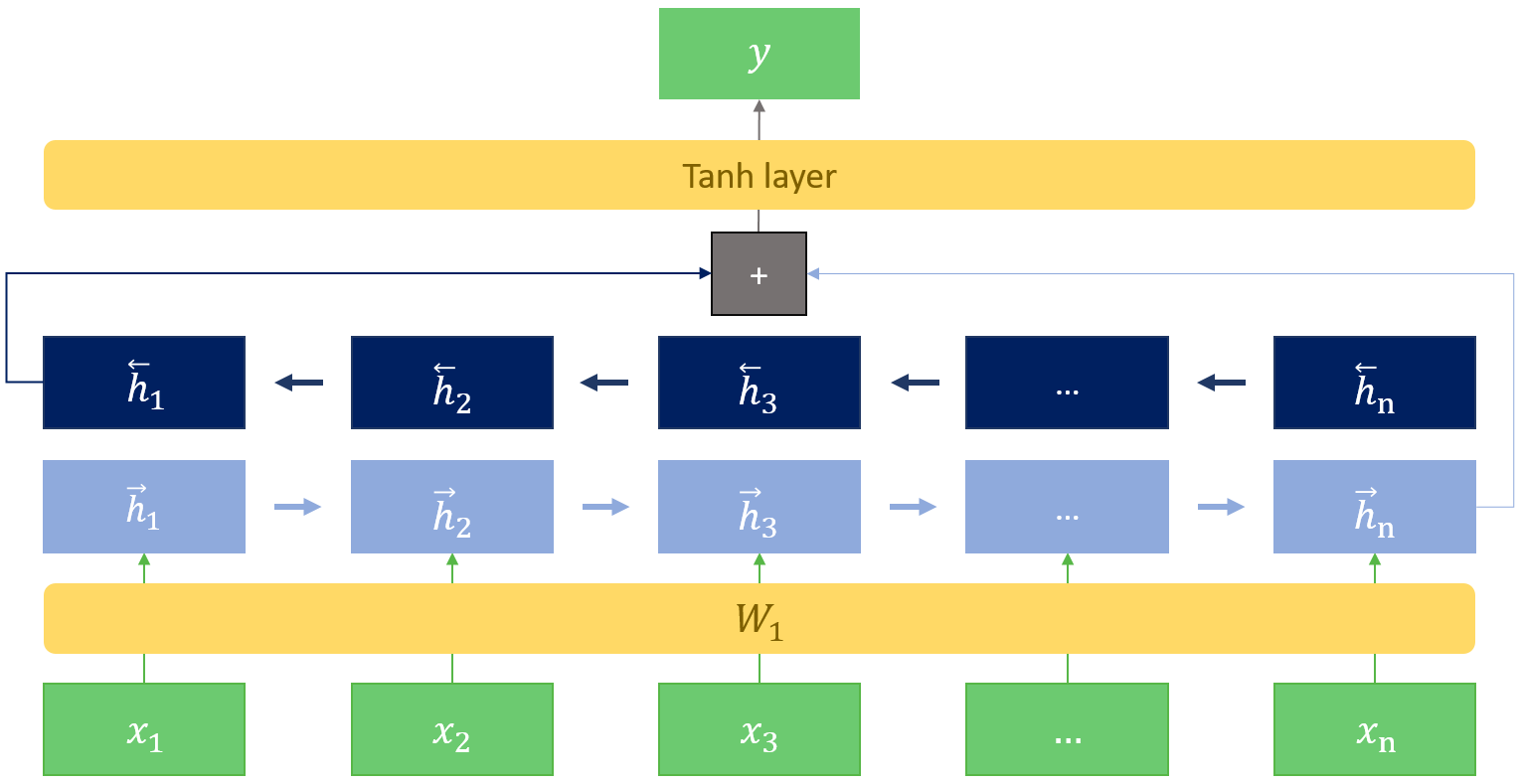}
\caption{Diagram of RNN model with bidirectional LSTM.
This diagram is identical to Figure \ref{hill_recurrent}, except the arrow above the LSTM state ($h_i$ ... $h_n$) represents LSTM direction. The outputs from the final state of the forward and backward LSTMs are concatenated together (as shown by the $+$ symbol) to form the final gloss representation vector which is fed through the Tanh layer.}
\label{bidirectional_diagram}
\end{figure}

As a further experiment, the single forward LSTM was replaced with a bidirectional LSTM. As discussed, bidirectional LSTMs are now somewhat standard in neural language modeling, as they reduce the effects of vanishing gradients, similar to taking the LSTM average. Note that this was not implemented in conjunction with the LSTM average. A digram of this model can be seen in Figure \ref{bidirectional_diagram}.  

The results of training this model on the original definitions dataset without BPE segmentation can be found in Table \ref{bidirectional_comparison}.

\begin{table}[!htpb]
\caption{Comparison of gloss encoder architectures}
\centering
\begin{tabular}{lll}
\toprule
LSTM type     & Definitions median rank & Crossword median rank \\\midrule
Single        & 106.5              & -                \\
Average       & 72.5               & \textbf{304.5}   \\
Bidirectional & \textbf{69.5}      & 380             
\end{tabular}
\label{bidirectional_comparison}

All models were trained on the definitions dataset for 10 epochs, crossword median rank results unavailable for single forward LSTM.
\end{table}

The model trained with a bidirectional LSTM achieved a median rank on the concept descriptions test set of 69.5, which is an improvement over the single forward LSTM. This is also slightly better than the LSTM average model on the concept descriptions test set. For optimum performance on the reverse dictionary task therefore, it appears that the bidirectional LSTM is the best model because it creates a more balanced representation of longer sequences. The performance of the bidirectional model was not as good as the LSTM average on the crossword set, mostly due to poor performance on shorter questions. Given time constraints, the bidirectional model could not be trained on all datasets, therefore the models reported in Tables \ref{overall_results} and \ref{crossword_results} were all trained using LSTM average.

\subsection{Discussion of Extensions}
Both of the extensions made to the gloss encoder LSTM proved to have a positive effect on model performance, as shown in Table \ref{bidirectional_comparison}. Because all 8 models could not be trained with both types of encoders, the LSTM average was chosen due to superior performance on the crossword test sets. It would be interesting to train more models with a bidirectional LSTM in future work.

With reference to Table \ref{overall_results}, it can be observed that all of the extensions served to improve results on the concept descriptions test set, with the best median rank and percentage of correct answers in top 10 model outputs on this test set were attained by the RNN BPE cosine model trained on the full dataset. 

The results on the crossword test sets were almost all best on the word-level models, but this dissertation suggests that this is due to overtraining for the BPE models, and that if these had not been trained for as long, they would have performed better. It is without question that including crossword questions in the training data resulted in considerable improvement across all models, as shown in Table \ref{crossword_results}. 

\section{Future extension: Target side sub-word units}
\begin{figure}[!htpb]
\centering
\includegraphics[width=1\linewidth]{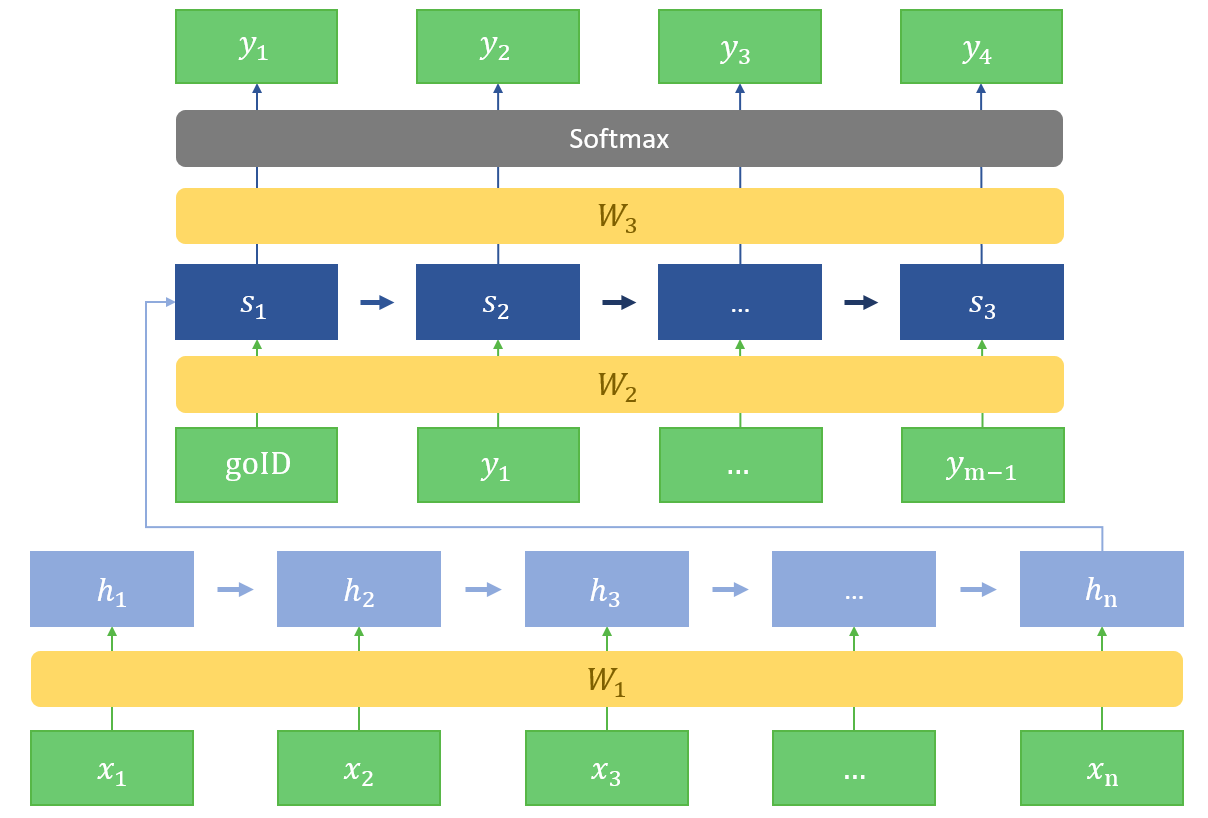}
\caption{Diagram of encoder-decoder model.
$x_i$ … $x_n$ are the inputs, the integer represents the timestep $t$, $n$ is the length of the sequence, $h_i$ … $h_n$ represent the encoder LSTM states at each timestep. In this diagram the final LSTM state is used as the input to the decoder, but any of the other methods (average, bidirectional LSTM) could also be used as input. The decoder is also an RNN which receives the encoder output and a special start symbol as input at $t=1$, then at each subsequent timestep it takes as input the output from the previous RNN unit and the output word at time $t-1$.}
\label{decoder_model}
\end{figure}

Thus far in this dissertation, experimentation with sub-word units has been limited to the source side (glosses). This is because the initial model did not include a decoder to allow for multiple-word or sub-word unit output. This is also the reason why multiple-word crossword questions had to be removed from the training set. Encoder-decoder models are commonplace in neural machine translation, meaning it would be quite possible to extend this model with a decoder. An example diagram of the model extended with a decoder is shown in Fig \ref{decoder_model}.

It would be desirable to allow sub-word outputs for multiple reasons:
\begin{itemize}
\item[1.] To allow for an unrestricted output vocabulary. Currently the model output is limited to only the number of words in the pretrained embeddings file. Embeddings files are available with over 1 million unique words, but these are extremely large files and loading them becomes costly.
\item[2.] To allow outputs of multiple words. Adding a decoder to the model would be beneficial even for word-level models to answer multiple-word crossword answers.
\item[3.] To learn more granular connections between glosses and heads. Using a sub-word model such as BPE may allow the model to learn that certain descriptions pertain to certain sub-word units or certain morphemes. 
\end{itemize}

As a baseline implementation, a sequence to sequence model is trained using Nematus \cite{sennrich-EtAl:2017:EACLDemo}. The training data used is the full training set which includes both definition and crossword data, and the test sets are the same five test sets used in other experiments. Because the head words are all singular words in the training and test sets, the outputs will be character-level. This is because applying BPE segmentation to the head words results in most of the heads staying as complete words, and would therefore not give great insight into the model performance. 

These models could not be run on the available GPUs due to software incompatibility errors. Due to these hardware constraints, the size of the learned representations and hidden dimensions had to be reduced to 64 and 128 respectively, and the batch size was doubled to 160 to allow reasonable training time. All other hyperparameters remained the same.

Quantitative results are not available due to the formatting of the test data, and the evaluation script reports a BLEU score, which may not be appropriate for a character-level evaluation.

Qualitative analysis of some training outputs can still be conducted, as the model intermittently outputs samples of training hypotheses. Three sample outputs after training for five epochs are included in Table \ref{5_epoch_char}. The training data used was not cleaned to remove the head words from the glosses before training, which has resulted in the model learning to reproduce gloss words as the lowest cost output words. For the first head word \textit{inert}, the head word is included in the definition, so some of the ouptuts are quite close: \textit{inertia}, \textit{inative}, \textit{inert}. This behaviour becomes more clear in the second example, the head word is not included in the definition, so the model outputs words that closely resemble one of the gloss words \textit{commencing}; eg: \textit{commerciator}, \textit{commons}, \textit{commerce} etc. One very notable feature of all of the sample outputs after 5 epochs is that they are real words, which is not a guarantee of a character-level model. This is exemplified in Table \ref{6_epoch_char}, showing sample outputs after an additional epoch of training. Here all words in bold are not real words. This is one undesirable consequence of having an unlimited vocabulary. Some of the words however are promising examples of why we wish to have sub-word output, for example one output for head word \textit{homicide} is \textit{premedicide}. While not a real word, it is a good example of how appropriate suffixes can be learned and applied in novel contexts. 

\begin{table}[!htbp]
\caption{Nematus character level outputs after 5 epochs}
\begin{tabular}{p{8cm} l l}
\toprule
Gloss                                                                                            & Head              & Output hypotheses       \\\midrule
in chemistry the term inert is used to describe something that is not chemically active          & i n e r t         & i n e r t i a           \\
                                                                                                 &                   & n e u t r a l i s m     \\
                                                                                                 &                   & n e u t r a l           \\
                                                                                                 &                   & i n n a t i o n         \\
                                                                                                 &                   & i n n a t e             \\
                                                                                                 &                   & i n e r t               \\
                                                                                                 &                   & r a n g e d             \\
                                                                                                 &                   & r a n g e               \\
                                                                                                 &                   & i n n o c e n c e       \\
                                                                                                 &                   & n e r v e s             \\
                                                                                                 &                   & n e r v e               \\
                                                                                                 &                   & i n n i n g             \\
the twelfth month of the french republican calendar comm@@ en@@ cing august and ending september & f r u c t i d o r & c o m m e r c i a t o r \\
                                                                                                 &                   & c o m m o n s           \\
                                                                                                 &                   & c o m m e r c e         \\
                                                                                                 &                   & c o m m e n c e         \\
                                                                                                 &                   & c o m m e r c i a       \\
                                                                                                 &                   & c o m m u n e           \\
                                                                                                 &                   & c o m m u n i s m       \\
                                                                                                 &                   & f r u i t               \\
                                                                                                 &                   & c h r o m e             \\
                                                                                                 &                   & c o m m u n a           \\
                                                                                                 &                   & c o m b                 \\
                                                                                                 &                   & c o m m o n             \\
someone who is so ar@@ d@@ ently devoted to something that it resembles an addiction             & a d d i c t       & a n t i c               \\
                                                                                                 &                   & a n g e l               \\
                                                                                                 &                   & a n g e r               \\
                                                                                                 &                   & d e a d                 \\
                                                                                                 &                   & a d d i c t e r         \\
                                                                                                 &                   & a d d r e s s i o n     \\
                                                                                                 &                   & a d d i c t             \\
                                                                                                 &                   & a d d i c t o r         \\
                                                                                                 &                   & a d d e n d             \\
                                                                                                 &                   & a d d r e s s           \\
                                                                                                 &                   & a d d i c t i o n       \\
                                                                                                 &                   & a d d i c t e d        
\end{tabular}
\label{5_epoch_char}
\end{table}

\begin{table}[!htpb]
\caption{Nematus character level outputs after 6 epochs}
\begin{tabular}{p{8cm} l p{5cm}}
\toprule
Gloss                                                                            & Head            & Output hypotheses                                                                                                                                                                                                                                                                                                                                                                                               \\\midrule
in plate glass manufacturing the front door of the an@@ ne@@ aling arch          & t h r o a t     & a r c h i t e c t u r e                                                                                                                                                                                                                                                                                                                                                                                         \\
                                                                                 &                 & p l a t e s                                                                                                                                                                                                                                                                                                                                                                                                     \\
                                                                                 &                 & b l a n c h                                                                                                                                                                                                                                                                                                                                                                                                     \\
                                                                                 &                 & a r c h e r                                                                                                                                                                                                                                                                                                                                                                                                     \\
                                                                                 &                 & g l a n c e                                                                                                                                                                                                                                                                                                                                                                                                     \\
                                                                                 &                 & g l a d d e n                                                                                                                                                                                                                                                                                                                                                                                                   \\
                                                                                 &                 & p l a t e r                                                                                                                                                                                                                                                                                                                                                                                                     \\
                                                                                 &                 & p l a n t                                                                                                                                                                                                                                                                                                                                                                                                       \\
                                                                                 &                 & p l a t                                                                                                                                                                                                                                                                                                                                                                                                         \\
                                                                                 &                 & g l a s s                                                                                                                                                                                                                                                                                                                                                                                                       \\
                                                                                 &                 & a r c h                                                                                                                                                                                                                                                                                                                                                                                                         \\
                                                                                 &                 & p l a t e                                                                                                                                                                                                                                                                                                                                                                                                       \\
sports the weight a horse must carry in a han@@ dic@@ ap race                    & i m p o s t     & \textbf{h a n d i c i d i t y}                                                                                                                                                                                                                                                                                                                                                                                           \\
                                                                                 &                 & p r o g r e s s                                                                                                                                                                                                                                                                                                                                                                                                 \\
                                                                                 &                 & \textbf{h a n d i c a t e}                                                                                                                                                                                                                                                                                                                                                                                               \\
                                                                                 &                 & p r o g r e s s i o n                                                                                                                                                                                                                                                                                                                                                                                           \\
                                                                                 &                 & \textbf{h a n d i e}                                                                                                                                                                                                                                                                                                                                                                                                     \\
                                                                                 &                 & p l a y                                                                                                                                                                                                                                                                                                                                                                                                         \\
                                                                                 &                 & h a n d i c a p                                                                                                                                                                                                                                                                                                                                                                                                 \\
                                                                                 &                 & p o r t                                                                                                                                                                                                                                                                                                                                                                                                         \\
                                                                                 &                 & h a n g                                                                                                                                                                                                                                                                                                                                                                                                         \\
                                                                                 &                 & \textbf{h a n d h e a d}                                                                                                                                                                                                                                                                                                                                                                                                 \\
                                                                                 &                 & h a n d                                                                                                                                                                                                                                                                                                                                                                                                         \\
                                                                                 &                 & h a n d i n g                                                                                                                                                                                                                                                                                                                                                                                                   \\
the killing of one person by another whether pre@@ med@@ itated or unintentional & h o m i c i d e & \textbf{p r e d i c a t i o n a l i s m}                                                                                                                                                                                                                                                                                                                                                                                 \\
                                                                                 &                 & \textbf{p r e m e d i c}                                                                                                                                                                                                                                                                                                                                                                                                 \\
                                                                                 &                 & \textbf{p r e m e d i c i d e}                                                                                                                                                                                                                                                                                                                                                                                           \\
                                                                                 &                 & \textbf{p r e m e d i c i c}                                                                                                                                                                                                                                                                                                                                                                                             \\
                                                                                 &                 & m e d i a t i v e                                                                                                                                                                                                                                                                                                                                                                                               \\
                                                                                 &                 & c o n d e m n i n g                                                                                                                                                                                                                                                                                                                                                                                             \\
                                                                                 &                 & p r e d i c a t i o n                                                                                                                                                                                                                                                                                                                                                                                           \\
                                                                                 &                 & \textbf{p r e m i e r y}                                                                                                                                                                                                                                                                                                                                                                                                 \\
                                                                                 &                 & c o n d i t i o n i n g                                                                                                                                                                                                                                                                                                                                                                                         \\
                                                                                 &                 & p r e m i e r                                                                                                                                                                                                                                                                                                                                                                                                   \\
                                                                                 &                 & m e d i a n                                                                                                                                                                                                                                                                                                                                                                                                     \\
                                                                                 &                 & \textbf{p r e d i c a t i o n a l i c i c i c i d i c i d i c i d i c i d i c i d i c i d i c i d i c i d i c i d i c i d i c i d i c i d i c i d i c i d i c i d i c (...)}
\end{tabular}
\label{6_epoch_char}
\end{table}

\newpage

\section{Conclusion}
In the interest of furthering research into definition models, this dissertation evaluates the work of Hill et al. \citeyear{hill2015learning} and identifies some areas in which their definition models could be extended, specifically those using RNNs. Inspiration for the extensions is drawn from recent research in neural language modeling and neural machine translation. The main goal is to improve model outputs for general knowledge crossword questions. Hill et al. \citeyear{hill2015learning} do not include crossword questions in their training data, so the dataset is extended to include over 350,000 crossword questions. Experiments are also carried out on a BPE-segmented training set \cite{sennrich2015neural}. Further extensions are made to the model itself, including a change to the gloss padding algorithm to allow for training on longer sequences, and two changes to the RNN used as a gloss encoder, namely taking an average of LSTM hidden states across the sequence and using a bidirectional LSTM. These extensions all result in an improved performance versus the base model, with the most significant improvement attained by including crossword data in the training set. 

Some preliminary experimentation is also carried out by training a sequence-to-sequence model using Nematus \cite{sennrich-EtAl:2017:EACLDemo} with BPE-segmented glosses and character-separated head words. While no insight can be gained into the outright performance of the model, by qualitatively analysing some sample training outputs, it is clear that the model was able to output the correct word in some cases, and that it is showing some promising signs by applying affixes correctly in novel contexts.

Finally some recommendations for future work are suggested. Research could be carried out on fine-tuning fully-trained models on just the crossword dataset to see if crossword performance improves, as the combined dataset still consists of mostly dictionary definitions, which may be negatively affecting model performance on the crossword test examples. All hyperparameters could be more thoroughly investigated through grid search to find the optimum parameters, including changes to the number of BPE merge operations in sub-word experiments. Lastly, it would be interesting to continue experimentation on sequence-to-sequence models by adding a decoder to the model, as shown in Figure \ref{decoder_model}, to allow character-level and/or multiple-word output, as well as an unlimited output vocabulary.

\newpage
\bibliographystyle{apacite}
\bibliography{bibliography}

\begin{thebibliography}{}

\bibitem [\protect \citeauthoryear {%
Abadi%
\ \protect \BOthers {.}}{%
Abadi%
\ \protect \BOthers {.}}{%
{\protect \APACyear {2016}}%
}]{%
abadi2016tensorflow}
\APACinsertmetastar {%
abadi2016tensorflow}%
\begin{APACrefauthors}%
Abadi, M.%
, Barham, P.%
, Chen, J.%
, Chen, Z.%
, Davis, A.%
, Dean, J.%
\BDBL {}others%
\end{APACrefauthors}%
\unskip\
\newblock
\APACrefYearMonthDay{2016}{}{}.
\newblock
{\BBOQ}\APACrefatitle {Tensorflow: a system for large-scale machine learning.}
  {Tensorflow: a system for large-scale machine learning.}{\BBCQ}
\newblock
\BIn{} \APACrefbtitle {OSDI} {Osdi}\ (\BVOL~16, \BPGS\ 265--283).
\PrintBackRefs{\CurrentBib}

\bibitem [\protect \citeauthoryear {%
Bachrach%
\ \protect \BOthers {.}}{%
Bachrach%
\ \protect \BOthers {.}}{%
{\protect \APACyear {2017}}%
}]{%
bachrach2017attention}
\APACinsertmetastar {%
bachrach2017attention}%
\begin{APACrefauthors}%
Bachrach, Y.%
, Zukov-Gregoric, A.%
, Coope, S.%
, Tovell, E.%
, Maksak, B.%
, Rodriguez, J.%
\BDBL {}Bordbar, M.%
\end{APACrefauthors}%
\unskip\
\newblock
\APACrefYearMonthDay{2017}{}{}.
\newblock
{\BBOQ}\APACrefatitle {An Attention Mechanism for Neural Answer Selection Using
  a Combined Global and Local View} {An attention mechanism for neural answer
  selection using a combined global and local view}.{\BBCQ}
\newblock
\BIn{} \APACrefbtitle {Tools with Artificial Intelligence (ICTAI), 2017 IEEE
  29th International Conference on} {Tools with artificial intelligence
  (ictai), 2017 ieee 29th international conference on}\ (\BPGS\ 425--432).
\PrintBackRefs{\CurrentBib}

\bibitem [\protect \citeauthoryear {%
Bengio%
, Ducharme%
, Vincent%
\BCBL {}\ \BBA {} Jauvin%
}{%
Bengio%
\ \protect \BOthers {.}}{%
{\protect \APACyear {2003}}%
}]{%
bengio2003neural}
\APACinsertmetastar {%
bengio2003neural}%
\begin{APACrefauthors}%
Bengio, Y.%
, Ducharme, R.%
, Vincent, P.%
\BCBL {}\ \BBA {} Jauvin, C.%
\end{APACrefauthors}%
\unskip\
\newblock
\APACrefYearMonthDay{2003}{}{}.
\newblock
{\BBOQ}\APACrefatitle {A neural probabilistic language model} {A neural
  probabilistic language model}.{\BBCQ}
\newblock
\APACjournalVolNumPages{Journal of machine learning
  research}{3}{Feb}{1137--1155}.
\PrintBackRefs{\CurrentBib}

\bibitem [\protect \citeauthoryear {%
Bergstra%
\ \protect \BOthers {.}}{%
Bergstra%
\ \protect \BOthers {.}}{%
{\protect \APACyear {2010}}%
}]{%
bergstra2010theano}
\APACinsertmetastar {%
bergstra2010theano}%
\begin{APACrefauthors}%
Bergstra, J.%
, Breuleux, O.%
, Bastien, F.%
, Lamblin, P.%
, Pascanu, R.%
, Desjardins, G.%
\BDBL {}Bengio, Y.%
\end{APACrefauthors}%
\unskip\
\newblock
\APACrefYearMonthDay{2010}{}{}.
\newblock
{\BBOQ}\APACrefatitle {Theano: A CPU and GPU math compiler in Python} {Theano:
  A cpu and gpu math compiler in python}.{\BBCQ}
\newblock
\BIn{} \APACrefbtitle {Proc. 9th Python in Science Conf} {Proc. 9th python in
  science conf}\ (\BVOL~1).
\PrintBackRefs{\CurrentBib}

\bibitem [\protect \citeauthoryear {%
Bilac%
, Watanabe%
, Hashimoto%
, Tokunaga%
\BCBL {}\ \BBA {} Tanaka%
}{%
Bilac%
\ \protect \BOthers {.}}{%
{\protect \APACyear {2004}}%
}]{%
bilac2004dictionary}
\APACinsertmetastar {%
bilac2004dictionary}%
\begin{APACrefauthors}%
Bilac, S.%
, Watanabe, W.%
, Hashimoto, T.%
, Tokunaga, T.%
\BCBL {}\ \BBA {} Tanaka, H.%
\end{APACrefauthors}%
\unskip\
\newblock
\APACrefYearMonthDay{2004}{}{}.
\newblock
{\BBOQ}\APACrefatitle {Dictionary search based on the target word description}
  {Dictionary search based on the target word description}.{\BBCQ}
\newblock
\BIn{} \APACrefbtitle {Proc. of the Tenth Annual Meeting of The Association for
  Natural Language Processing (NLP2004)} {Proc. of the tenth annual meeting of
  the association for natural language processing (nlp2004)}\ (\BPGS\
  556--559).
\PrintBackRefs{\CurrentBib}

\bibitem [\protect \citeauthoryear {%
Hill%
, Cho%
, Korhonen%
\BCBL {}\ \BBA {} Bengio%
}{%
Hill%
\ \protect \BOthers {.}}{%
{\protect \APACyear {2015}}%
}]{%
hill2015learning}
\APACinsertmetastar {%
hill2015learning}%
\begin{APACrefauthors}%
Hill, F.%
, Cho, K.%
, Korhonen, A.%
\BCBL {}\ \BBA {} Bengio, Y.%
\end{APACrefauthors}%
\unskip\
\newblock
\APACrefYearMonthDay{2015}{}{}.
\newblock
{\BBOQ}\APACrefatitle {Learning to understand phrases by embedding the
  dictionary} {Learning to understand phrases by embedding the
  dictionary}.{\BBCQ}
\newblock
\APACjournalVolNumPages{arXiv preprint arXiv:1504.00548}{}{}{}.
\PrintBackRefs{\CurrentBib}

\bibitem [\protect \citeauthoryear {%
Huang%
, Socher%
, Manning%
\BCBL {}\ \BBA {} Ng%
}{%
Huang%
\ \protect \BOthers {.}}{%
{\protect \APACyear {2012}}%
}]{%
huang2012improving}
\APACinsertmetastar {%
huang2012improving}%
\begin{APACrefauthors}%
Huang, E\BPBI H.%
, Socher, R.%
, Manning, C\BPBI D.%
\BCBL {}\ \BBA {} Ng, A\BPBI Y.%
\end{APACrefauthors}%
\unskip\
\newblock
\APACrefYearMonthDay{2012}{}{}.
\newblock
{\BBOQ}\APACrefatitle {Improving word representations via global context and
  multiple word prototypes} {Improving word representations via global context
  and multiple word prototypes}.{\BBCQ}
\newblock
\BIn{} \APACrefbtitle {Proceedings of the 50th Annual Meeting of the
  Association for Computational Linguistics: Long Papers-Volume 1} {Proceedings
  of the 50th annual meeting of the association for computational linguistics:
  Long papers-volume 1}\ (\BPGS\ 873--882).
\PrintBackRefs{\CurrentBib}

\bibitem [\protect \citeauthoryear {%
Koehn%
}{%
Koehn%
}{%
{\protect \APACyear {2017}}%
}]{%
DBLP:journals/corr/abs-1709-07809}
\APACinsertmetastar {%
DBLP:journals/corr/abs-1709-07809}%
\begin{APACrefauthors}%
Koehn, P.%
\end{APACrefauthors}%
\unskip\
\newblock
\APACrefYearMonthDay{2017}{}{}.
\newblock
{\BBOQ}\APACrefatitle {Neural Machine Translation} {Neural machine
  translation}.{\BBCQ}
\newblock
\APACjournalVolNumPages{CoRR}{abs/1709.07809}{}{}.
\newblock
\begin{APACrefURL} \url{http://arxiv.org/abs/1709.07809} \end{APACrefURL}
\PrintBackRefs{\CurrentBib}

\bibitem [\protect \citeauthoryear {%
Lee%
, Cho%
\BCBL {}\ \BBA {} Hofmann%
}{%
Lee%
\ \protect \BOthers {.}}{%
{\protect \APACyear {2016}}%
}]{%
lee2016fully}
\APACinsertmetastar {%
lee2016fully}%
\begin{APACrefauthors}%
Lee, J.%
, Cho, K.%
\BCBL {}\ \BBA {} Hofmann, T.%
\end{APACrefauthors}%
\unskip\
\newblock
\APACrefYearMonthDay{2016}{}{}.
\newblock
{\BBOQ}\APACrefatitle {Fully character-level neural machine translation without
  explicit segmentation} {Fully character-level neural machine translation
  without explicit segmentation}.{\BBCQ}
\newblock
\APACjournalVolNumPages{arXiv preprint arXiv:1610.03017}{}{}{}.
\PrintBackRefs{\CurrentBib}

\bibitem [\protect \citeauthoryear {%
Luong%
\ \BBA {} Manning%
}{%
Luong%
\ \BBA {} Manning%
}{%
{\protect \APACyear {2016}}%
}]{%
luong2016achieving}
\APACinsertmetastar {%
luong2016achieving}%
\begin{APACrefauthors}%
Luong, M\BHBI T.%
\BCBT {}\ \BBA {} Manning, C\BPBI D.%
\end{APACrefauthors}%
\unskip\
\newblock
\APACrefYearMonthDay{2016}{}{}.
\newblock
{\BBOQ}\APACrefatitle {Achieving open vocabulary neural machine translation
  with hybrid word-character models} {Achieving open vocabulary neural machine
  translation with hybrid word-character models}.{\BBCQ}
\newblock
\APACjournalVolNumPages{arXiv preprint arXiv:1604.00788}{}{}{}.
\PrintBackRefs{\CurrentBib}

\bibitem [\protect \citeauthoryear {%
Mach{\'a}{\v{c}}ek%
, Vidra%
\BCBL {}\ \BBA {} Bojar%
}{%
Mach{\'a}{\v{c}}ek%
\ \protect \BOthers {.}}{%
{\protect \APACyear {2018}}%
}]{%
machavcek2018morphological}
\APACinsertmetastar {%
machavcek2018morphological}%
\begin{APACrefauthors}%
Mach{\'a}{\v{c}}ek, D.%
, Vidra, J.%
\BCBL {}\ \BBA {} Bojar, O.%
\end{APACrefauthors}%
\unskip\
\newblock
\APACrefYearMonthDay{2018}{}{}.
\newblock
{\BBOQ}\APACrefatitle {Morphological and Language-Agnostic Word Segmentation
  for NMT} {Morphological and language-agnostic word segmentation for
  nmt}.{\BBCQ}
\newblock
\APACjournalVolNumPages{arXiv preprint arXiv:1806.05482}{}{}{}.
\PrintBackRefs{\CurrentBib}

\bibitem [\protect \citeauthoryear {%
Mikolov%
, Chen%
, Corrado%
\BCBL {}\ \BBA {} Dean%
}{%
Mikolov%
\ \protect \BOthers {.}}{%
{\protect \APACyear {2013}}%
}]{%
mikolov2013efficient}
\APACinsertmetastar {%
mikolov2013efficient}%
\begin{APACrefauthors}%
Mikolov, T.%
, Chen, K.%
, Corrado, G.%
\BCBL {}\ \BBA {} Dean, J.%
\end{APACrefauthors}%
\unskip\
\newblock
\APACrefYearMonthDay{2013}{}{}.
\newblock
{\BBOQ}\APACrefatitle {Efficient estimation of word representations in vector
  space} {Efficient estimation of word representations in vector space}.{\BBCQ}
\newblock
\APACjournalVolNumPages{arXiv preprint arXiv:1301.3781}{}{}{}.
\PrintBackRefs{\CurrentBib}

\bibitem [\protect \citeauthoryear {%
Mitchell%
\ \BBA {} Lapata%
}{%
Mitchell%
\ \BBA {} Lapata%
}{%
{\protect \APACyear {2010}}%
}]{%
mitchell2010composition}
\APACinsertmetastar {%
mitchell2010composition}%
\begin{APACrefauthors}%
Mitchell, J.%
\BCBT {}\ \BBA {} Lapata, M.%
\end{APACrefauthors}%
\unskip\
\newblock
\APACrefYearMonthDay{2010}{}{}.
\newblock
{\BBOQ}\APACrefatitle {Composition in distributional models of semantics}
  {Composition in distributional models of semantics}.{\BBCQ}
\newblock
\APACjournalVolNumPages{Cognitive science}{34}{8}{1388--1429}.
\PrintBackRefs{\CurrentBib}

\bibitem [\protect \citeauthoryear {%
Och%
, Tillmann%
\BCBL {}\ \BBA {} Ney%
}{%
Och%
\ \protect \BOthers {.}}{%
{\protect \APACyear {1999}}%
}]{%
och1999improved}
\APACinsertmetastar {%
och1999improved}%
\begin{APACrefauthors}%
Och, F\BPBI J.%
, Tillmann, C.%
\BCBL {}\ \BBA {} Ney, H.%
\end{APACrefauthors}%
\unskip\
\newblock
\APACrefYearMonthDay{1999}{}{}.
\newblock
{\BBOQ}\APACrefatitle {Improved alignment models for statistical machine
  translation} {Improved alignment models for statistical machine
  translation}.{\BBCQ}
\newblock
\BIn{} \APACrefbtitle {1999 Joint SIGDAT Conference on Empirical Methods in
  Natural Language Processing and Very Large Corpora.} {1999 joint sigdat
  conference on empirical methods in natural language processing and very large
  corpora.}
\PrintBackRefs{\CurrentBib}

\bibitem [\protect \citeauthoryear {%
Pennington%
, Socher%
\BCBL {}\ \BBA {} Manning%
}{%
Pennington%
\ \protect \BOthers {.}}{%
{\protect \APACyear {2014}}%
}]{%
pennington2014glove}
\APACinsertmetastar {%
pennington2014glove}%
\begin{APACrefauthors}%
Pennington, J.%
, Socher, R.%
\BCBL {}\ \BBA {} Manning, C.%
\end{APACrefauthors}%
\unskip\
\newblock
\APACrefYearMonthDay{2014}{}{}.
\newblock
{\BBOQ}\APACrefatitle {Glove: Global vectors for word representation} {Glove:
  Global vectors for word representation}.{\BBCQ}
\newblock
\BIn{} \APACrefbtitle {Proceedings of the 2014 conference on empirical methods
  in natural language processing (EMNLP)} {Proceedings of the 2014 conference
  on empirical methods in natural language processing (emnlp)}\ (\BPGS\
  1532--1543).
\PrintBackRefs{\CurrentBib}

\bibitem [\protect \citeauthoryear {%
Pereira%
, Tishby%
\BCBL {}\ \BBA {} Lee%
}{%
Pereira%
\ \protect \BOthers {.}}{%
{\protect \APACyear {1993}}%
}]{%
Pereira:1993:DCE:981574.981598}
\APACinsertmetastar {%
Pereira:1993:DCE:981574.981598}%
\begin{APACrefauthors}%
Pereira, F.%
, Tishby, N.%
\BCBL {}\ \BBA {} Lee, L.%
\end{APACrefauthors}%
\unskip\
\newblock
\APACrefYearMonthDay{1993}{}{}.
\newblock
{\BBOQ}\APACrefatitle {Distributional Clustering of English Words}
  {Distributional clustering of english words}.{\BBCQ}
\newblock
\BIn{} \APACrefbtitle {Proceedings of the 31st Annual Meeting on Association
  for Computational Linguistics} {Proceedings of the 31st annual meeting on
  association for computational linguistics}\ (\BPGS\ 183--190).
\newblock
\APACaddressPublisher{Stroudsburg, PA, USA}{Association for Computational
  Linguistics}.
\newblock
\begin{APACrefURL} \url{https://doi.org/10.3115/981574.981598} \end{APACrefURL}
\newblock
\begin{APACrefDOI} \doi{10.3115/981574.981598} \end{APACrefDOI}
\PrintBackRefs{\CurrentBib}

\bibitem [\protect \citeauthoryear {%
Santoro%
\ \protect \BOthers {.}}{%
Santoro%
\ \protect \BOthers {.}}{%
{\protect \APACyear {2017}}%
}]{%
NIPS2017_7082}
\APACinsertmetastar {%
NIPS2017_7082}%
\begin{APACrefauthors}%
Santoro, A.%
, Raposo, D.%
, Barrett, D\BPBI G.%
, Malinowski, M.%
, Pascanu, R.%
, Battaglia, P.%
\BCBL {}\ \BBA {} Lillicrap, T.%
\end{APACrefauthors}%
\unskip\
\newblock
\APACrefYearMonthDay{2017}{}{}.
\newblock
{\BBOQ}\APACrefatitle {A simple neural network module for relational reasoning}
  {A simple neural network module for relational reasoning}.{\BBCQ}
\newblock
\BIn{} I.~Guyon\ \BOthers {.}\ (\BEDS), \APACrefbtitle {Advances in Neural
  Information Processing Systems 30} {Advances in neural information processing
  systems 30}\ (\BPGS\ 4967--4976).
\newblock
\APACaddressPublisher{}{Curran Associates, Inc.}
\newblock
\begin{APACrefURL}
  \url{http://papers.nips.cc/paper/7082-a-simple-neural-network-module-for-relational-reasoning.pdf}
  \end{APACrefURL}
\PrintBackRefs{\CurrentBib}

\bibitem [\protect \citeauthoryear {%
Sennrich%
\ \protect \BOthers {.}}{%
Sennrich%
\ \protect \BOthers {.}}{%
{\protect \APACyear {2017}}%
}]{%
sennrich-EtAl:2017:EACLDemo}
\APACinsertmetastar {%
sennrich-EtAl:2017:EACLDemo}%
\begin{APACrefauthors}%
Sennrich, R.%
, Firat, O.%
, Cho, K.%
, Birch, A.%
, Haddow, B.%
, Hitschler, J.%
\BDBL {}Nadejde, M.%
\end{APACrefauthors}%
\unskip\
\newblock
\APACrefYearMonthDay{2017}{April}{}.
\newblock
{\BBOQ}\APACrefatitle {Nematus: a Toolkit for Neural Machine Translation}
  {Nematus: a toolkit for neural machine translation}.{\BBCQ}
\newblock
\BIn{} \APACrefbtitle {Proceedings of the Software Demonstrations of the 15th
  Conference of the European Chapter of the Association for Computational
  Linguistics} {Proceedings of the software demonstrations of the 15th
  conference of the european chapter of the association for computational
  linguistics}\ (\BPGS\ 65--68).
\newblock
\APACaddressPublisher{Valencia, Spain}{Association for Computational
  Linguistics}.
\newblock
\begin{APACrefURL} \url{http://aclweb.org/anthology/E17-3017} \end{APACrefURL}
\PrintBackRefs{\CurrentBib}

\bibitem [\protect \citeauthoryear {%
Sennrich%
, Haddow%
\BCBL {}\ \BBA {} Birch%
}{%
Sennrich%
\ \protect \BOthers {.}}{%
{\protect \APACyear {2015}}%
}]{%
sennrich2015neural}
\APACinsertmetastar {%
sennrich2015neural}%
\begin{APACrefauthors}%
Sennrich, R.%
, Haddow, B.%
\BCBL {}\ \BBA {} Birch, A.%
\end{APACrefauthors}%
\unskip\
\newblock
\APACrefYearMonthDay{2015}{}{}.
\newblock
{\BBOQ}\APACrefatitle {Neural machine translation of rare words with subword
  units} {Neural machine translation of rare words with subword units}.{\BBCQ}
\newblock
\APACjournalVolNumPages{arXiv preprint arXiv:1508.07909}{}{}{}.
\PrintBackRefs{\CurrentBib}

\bibitem [\protect \citeauthoryear {%
Turian%
, Ratinov%
\BCBL {}\ \BBA {} Bengio%
}{%
Turian%
\ \protect \BOthers {.}}{%
{\protect \APACyear {2010}}%
}]{%
Turian:2010:WRS:1858681.1858721}
\APACinsertmetastar {%
Turian:2010:WRS:1858681.1858721}%
\begin{APACrefauthors}%
Turian, J.%
, Ratinov, L.%
\BCBL {}\ \BBA {} Bengio, Y.%
\end{APACrefauthors}%
\unskip\
\newblock
\APACrefYearMonthDay{2010}{}{}.
\newblock
{\BBOQ}\APACrefatitle {Word Representations: A Simple and General Method for
  Semi-supervised Learning} {Word representations: A simple and general method
  for semi-supervised learning}.{\BBCQ}
\newblock
\BIn{} \APACrefbtitle {Proceedings of the 48th Annual Meeting of the
  Association for Computational Linguistics} {Proceedings of the 48th annual
  meeting of the association for computational linguistics}\ (\BPGS\ 384--394).
\newblock
\APACaddressPublisher{Stroudsburg, PA, USA}{Association for Computational
  Linguistics}.
\newblock
\begin{APACrefURL} \url{http://dl.acm.org/citation.cfm?id=1858681.1858721}
  \end{APACrefURL}
\PrintBackRefs{\CurrentBib}

\bibitem [\protect \citeauthoryear {%
Virpioja%
, Smit%
, Gr{\"o}nroos%
, Kurimo%
\BCBL {}\ \protect \BOthers {.}}{%
Virpioja%
\ \protect \BOthers {.}}{%
{\protect \APACyear {2013}}%
}]{%
virpioja2013morfessor}
\APACinsertmetastar {%
virpioja2013morfessor}%
\begin{APACrefauthors}%
Virpioja, S.%
, Smit, P.%
, Gr{\"o}nroos, S\BHBI A.%
, Kurimo, M.%
\BCBL {}\ \BOthersPeriod {.}\end{APACrefauthors}%
\unskip\
\newblock
\APACrefYearMonthDay{2013}{}{}.
\newblock
{\BBOQ}\APACrefatitle {Morfessor 2.0: Python implementation and extensions for
  Morfessor Baseline} {Morfessor 2.0: Python implementation and extensions for
  morfessor baseline}.{\BBCQ}
\newblock

\PrintBackRefs{\CurrentBib}

\bibitem [\protect \citeauthoryear {%
Yin%
\ \protect \BOthers {.}}{%
Yin%
\ \protect \BOthers {.}}{%
{\protect \APACyear {2015}}%
}]{%
yin2015neural}
\APACinsertmetastar {%
yin2015neural}%
\begin{APACrefauthors}%
Yin, J.%
, Jiang, X.%
, Lu, Z.%
, Shang, L.%
, Li, H.%
\BCBL {}\ \BBA {} Li, X.%
\end{APACrefauthors}%
\unskip\
\newblock
\APACrefYearMonthDay{2015}{}{}.
\newblock
{\BBOQ}\APACrefatitle {Neural generative question answering} {Neural generative
  question answering}.{\BBCQ}
\newblock
\APACjournalVolNumPages{arXiv preprint arXiv:1512.01337}{}{}{}.
\PrintBackRefs{\CurrentBib}

\end{thebibliography}
\end{spacing}
\end{document}